\documentclass[conference]{IEEEtran}
\IEEEoverridecommandlockouts
\usepackage{cite}
\usepackage{amsmath,amssymb,amsfonts}
\usepackage{algorithmic}
\usepackage{graphicx}
\usepackage{textcomp}
\usepackage{xcolor}
\usepackage{subfigure}
\usepackage[export]{adjustbox}
\def\BibTeX{{\rm B\kern-.05em{\sc i\kern-.025em b}\kern-.08em
    T\kern-.1667em\lower.7ex\hbox{E}\kern-.125emX}}
\begin{document}
\bibliographystyle{ieeetr}

\title{
Crucial Semantic Classifier-based Adversarial Learning for Unsupervised Domain Adaptation}
\author{
    \IEEEauthorblockN{
        Yumin Zhang\IEEEauthorrefmark{1},
        Yajun Gao\IEEEauthorrefmark{2},
        Hongliu Li\IEEEauthorrefmark{3},
        Ating Yin\IEEEauthorrefmark{4},
        Duzhen Zhang\IEEEauthorrefmark{5},
        Xiuyi Chen\thanks{* Xiuyi Chen is the corresponding author (chenxiuyi01@baidu.com).}\IEEEauthorrefmark{1}
    }
    \IEEEauthorblockA{\IEEEauthorrefmark{1}Baidu Inc.,Beijing, 100080, China.}
    \IEEEauthorblockA{\IEEEauthorrefmark{2}Shenyang Institute of Automation, Chinese Academy of Sciences, Shenyang, 110016, China.}
    \IEEEauthorblockA{\IEEEauthorrefmark{3}University of Science and Technology of China, Hefei, 230026, China.}
    \IEEEauthorblockA{\IEEEauthorrefmark{4}College of Electrical and Information Engineering, Hunan University, Changsha, 410082, China.}
    \IEEEauthorblockA{\IEEEauthorrefmark{5}Institute of Automation, Chinese Academy of Sciences, Beijing, 100190, China.}
}

\maketitle

\begin{abstract}
Unsupervised Domain Adaptation (UDA), which aims to explore the transferrable features from a well-labeled source domain to a related unlabeled target domain, has been widely progressed. Nevertheless, as one of the mainstream, existing adversarial-based methods neglect to filter the irrelevant semantic knowledge, hindering adaptation performance improvement. Besides, they require an additional domain discriminator that strives extractor to generate confused representations, but discrete designing may cause model collapse. To tackle the above issues, we propose \underline{C}rucial \underline{S}emantic \underline{C}lassifier-based \underline{A}dversarial \underline{L}earning (CSCAL), which pays more attention to crucial semantic knowledge transferring and leverages the classifier to implicitly play the role of domain discriminator without extra network designing. Specifically, in intra-class-wise alignment, a Paired-Level Discrepancy (PLD) is designed to transfer crucial semantic knowledge. Additionally, based on classifier predictions, a Nuclear Norm-based Discrepancy (NND) is formed that considers inter-class-wise information and improves the adaptation performance. Moreover, CSCAL can be effortlessly merged into different UDA methods as a regularizer and dramatically promote their performance.
\end{abstract}


\section{Introduction}
Deep Neural Networks (DNNs) have shown impressive success in various computer vision tasks, e.g., image classification~\cite{he2016deep}, object detection~\cite{ren2015faster} and semantic segmentation~\cite{chen2014semantic,dong2020can}.
However, their success relies highly on massive well-labeled data, which is extremely labor-intensive and time-consuming.
Moreover, the appliance of DNNs is constrained by its poor generalization, due to the domain discrepancy between training data (a.k.a. source domain) and testing data (a.k.a. target domain), which makes the performance dramatically degenerate on new testing data.
Therefore, Unsupervised Domain Adaptation (UDA), which aims to transfer learned knowledge from a well-labeled source domain to a related but unlabeled target domain with different distribution~\cite{fang2020open}, has been deeply researched. 

\begin{figure}
    \centering
    \includegraphics[width=1.0\linewidth]{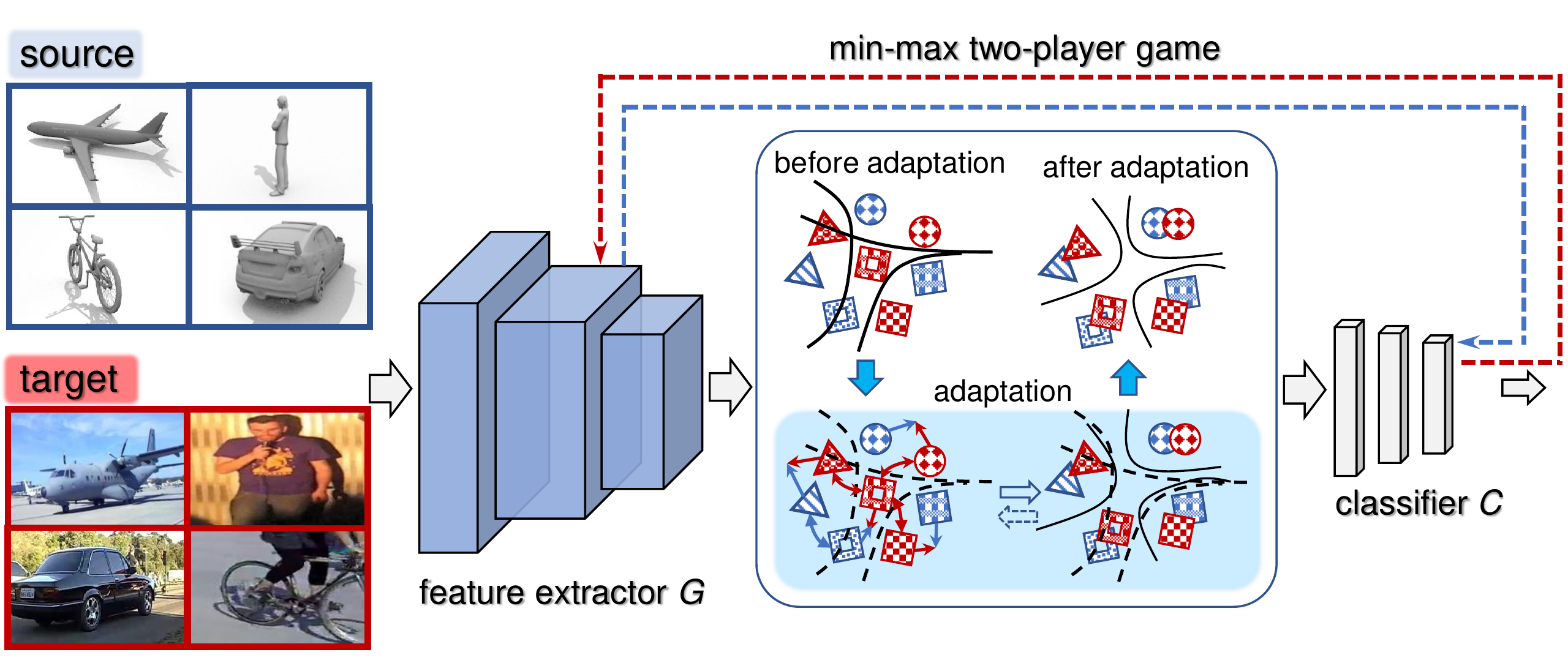}
    \caption{Illustration of our adversarial paradigm.
    Unlike popular adversarial methods that require a well-designed domain discriminator, we leverage the original classifier $C$ and feature extractor $G$ to play the min-max game, achieving domain-level alignment and category-level distinguishment simultaneously.}
    \label{fig:fig1}
\end{figure}

Generally speaking, we can organize existing methods into two mainstream, i.e., 
statistical discrepancy-based methods~\cite{li2020enhanced,long2015learning,long2018conditional} and adversarial-based methods~\cite{saito2018maximum,tzeng2017adversarial,li2021semantic}.
In statistical discrepancy-based methods, the domain discrepancy is minimized by reducing well-designed statistical discrepancy, such as Maximum Mean Discrepancy (MMD)~\cite{tzeng2014deep} and Joint Maximum Mean Discrepancy (JMMD)~\cite{long2017deep}.
In adversarial-based methods, a domain discriminator is designed to encourage domain-level feature alignment via an adversarial min-max two-player game.
Encouraged by the remarkable performance achieved by adversarial learning~\cite{chen2022reusing}, we developed our method based on the adversarial paradigm.

Although existing adversarial-based methods have achieved impressive performance,
nevertheless, most of them treat the crucial semantic knowledge and inessential semantic knowledge (e.g., the inevitable background information) as equivalent, causing negative transfer~\cite{li2021semantic,dong2021and}.
Moreover, these popular adversarial-based methods~\cite{zhang2019domain,lu2020stochastic,du2021cross} require an additional well-designed domain discriminator to align the cross-domain representations while neglecting the category-level information and may cause model collapse problems~\cite{tang2020discriminative}.


To address these problems, we propose a novel adversarial learning paradigm named \underline{C}rucial \underline{S}emantic \underline{C}lassifier-based \underline{A}dversarial \underline{L}earning (CSCAL), which consists of only a feature extractor $G$ and the classifier $C$.
As shown in Fig.~\ref{fig:fig1}, in our framework, in addition to predicting the category of the input image, classifier $C$ implicitly plays the role of domain discriminator in the min-max game.
In order to pay attention to the crucial semantic knowledge, inspired by the findings that the predictions of DNNs depend on the model concentrated discriminative region area~\cite{zhou2016learning}, the wrong predictions are leveraged to emphasize the crucial semantic knowledge to alleviate the negative transfer.
Specifically, for intra-class-wise alignment, we first define the data with the same labels as the paired samples including intra-domain and inter-domain paired samples. 
Then we construct the Paired-Level Discrepancy (PLD) and align those paired samples in a classifier-based adversarial manner.
We train the classifier to maximize the PLD that raises the weights of wrong predictions. Due to the adversarial mechanism, the feature extractor will strive to repress the corresponding inessential semantic knowledge and concentrate on the crucial semantic knowledge. 
Moreover, we take the inter-class-wise information into account to further improve adaptation performance. 
As the same classifier, the predictions from the source and target domain naturally present discrepancies due to the domain shift. 
Based on the prediction matrix calculated by the classifier, we construct a Nuclear Norm-based Discrepancy (NND) as domain critic, which further improves model adaptation.
Our main contributions in this paper are summarized as follows:
\begin{itemize}
    \item We propose a novel adversarial paradigm, CSCAL, which focuses on transferring crucial semantic knowledge without extra network design.
    \item Specifically, for intra-class-wise alignment, PLD is designed to discover and transfer crucial semantic features. 
    Moreover, considering inter-class-wise alignment, NND is formed to improve model adaptation.
    \item Without bells and whistles, CSCAL can seamlessly incorporate different UDA methods and significantly boost performance. Comprehensive experimental results and analysis on multiple UDA benchmarks, including Office-Home, DomainNet, and Office-31 demonstrate the effectiveness of CSCAL.
\end{itemize}
\section{Related Works}
\textbf{Adversarial Domain Adaptation.}
Inspired by the Generative Adversarial Network~\cite{goodfellow2020generative}, adversarial-based learning, playing a min-max game to get the domain-alignment knowledge, is a mainstream method in UDA.
Existing adversarial-based methods can be categorized into feature-level representations alignment~\cite{ganin2016domain,long2018conditional,gao2021gradient, tang2020discriminative} and image-to-image techniques~\cite{zhu2017unpaired,liu2017unsupervised,sankaranarayanan2018generate,hoffman2018cycada,pizzati2020domain}.
In the former, the domain discriminator was introduced in DNN~\cite{ganin2016domain} and CDAN~\cite{long2018conditional} to encourage the feature extractor to generate confusing features.
In the latter, the domain adaptation is achieved by transferring the raw source data to the target style~\cite{zhu2017unpaired,liu2017unsupervised,sankaranarayanan2018generate}. 

Our method lies in feature-level alignment. Unlike previous methods that require an additional domain discriminator to transfer knowledge, we reuse the original classifier to play the role of domain critic, which avoids the model collapse caused by distinct network design~\cite{tang2020discriminative}.


\textbf{Metric Matching Methods.}
These metric-based methods learn the transferable knowledge across domains by reducing the well-designed statistical discrepancy.
Among these methods, the Maximum Mean Discrepancy (MMD)~\cite{tzeng2014deep} and Joint Maximum Mean Discrepancy (JMMD)~\cite{long2017deep} have been widely used in various UDA tasks.
Besides, a weighted MMD was introduced in~\cite{yan2017mind} to alleviate class weight bias.
More recently, Margin Disparity Discrepancy~\cite{zhang2019bridging} was introduced to align the domain distributions with a rigorous generalization bound.
Besides, the Optimal Transport (OT) distance is also used for measuring the relationship between different distributions~\cite{arjovsky2017wasserstein}.
Aiming to tackle the bottleneck caused by a biased transport map, ETD~\cite{li2020enhanced} was proposed to achieve the feature alignment via an attention-aware transport distance.

Although the above methods have achieved remarkable advances in learning domain-invariant knowledge, they transfer the whole image features without processing the inessential semantic information, hindering further alignment.
In this paper, we concentrate on transferring crucial semantic knowledge to alleviate the negative transfer caused by inessential information.

\textbf{Attention-based Mechanism.}
The attention of a convolutional neural network is defined as a set of spatial maps that the network focuses on for performing a particular task~\cite{zagoruyko2016paying}, and attention-based mechanisms have achieved remarkable advances in deep learning~\cite{vaswani2017attention, chen2019all,yu2022metaformer}.
Some existing methods leverage different degrees of attention to boost the domain adaptation performance.
For example, AHT~\cite{moon2017completely} designed a heterogeneous transfer learning algorithm to transfer the necessary knowledge. 
Recently, TADA~\cite{wang2019transferable} leveraged the local and global attention mechanism to explore the transferable features across domains.
In domain generalization, Meng et al.~\cite{meng2022attention} introduced an attention diversification framework that reassigns proper attention to diverse task-related knowledge.

However, most of these attention-based methods require elaborately designed network architectures to obtain appropriate degrees of attention, constraining their realistic appliance. Unlike the methods mentioned above, we measure the discrepancy pair-wise on output space to discover and transfer crucial semantic knowledge. Besides, we also consider inter-class-wise divergence and construct NND as a domain critic to further domain adaptation.

\begin{figure*}
    \centering
    \includegraphics[width=1.0\linewidth]{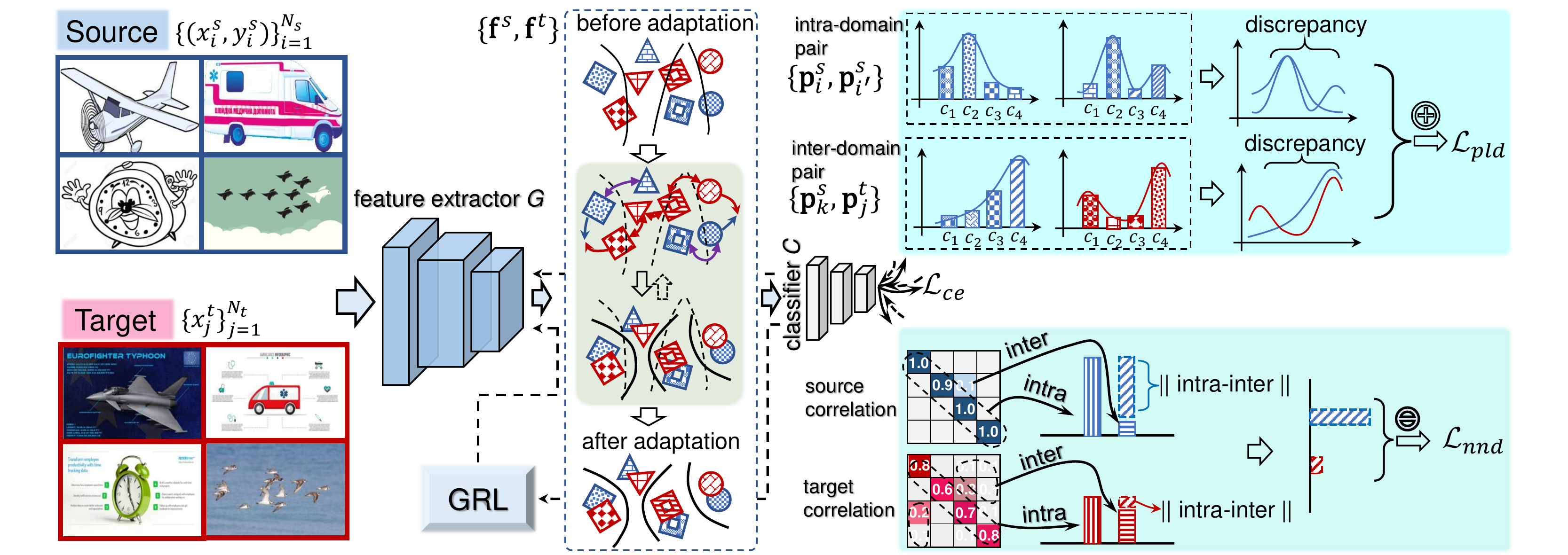}
    \caption{The conceptual overview of CSCAL consists of only a feature extractor $G$ and a classifier $C$. 
    $\mathcal{L}_{ce}$ is the cross-entropy loss on the source domain.
    $\mathcal{L}_{pld}$ and $\mathcal{L}_{nnd}$, based on the output of classifier $C$, are Paired-Level Discrepancy and Nuclear Norm-based Discrepancy, respectively. 
    With the help of $\mathcal{L}_{pld}$ and $\mathcal{L}_{nnd}$, classifier $C$ implicitly plays the role of discriminator in the min-max two-player game, and the Gradient Reverse Layer (GRL) is performed to achieve adversarial learning.}
    \label{fig:fig2}
\end{figure*}
\section{Method}
\subsection{Preliminary Knowledge and Overview}
\subsubsection{Preliminary Knowledge}
Given a source domain $\mathcal{D}_s$ and an unlabeled target domain $\mathcal{D}_t$, where $\mathcal{D}_s = \{(x_{i}^s, y_{i}^s)\}_{i=1}^{N_s}$ contains $N_s$ source samples $x_i^s$ and corresponding one-hot labels $y_i^s \in \{1, 2,..., K\}$, and $\mathcal{D}_t = \{x_j^t\}_{j=1}^{N_t}$ contains $N_t$ target samples $x_j^t$.
UDA aims to learn a function $\mathcal{F}$ that can predict reliable predictions on the target domain with only source-supervised information accessible.
Generally, traditional adversarial-based methods~\cite{ganin2016domain, long2018conditional, gao2021gradient, li2021bi, zhang2019domain} require the assistance of an additional well-designed discriminator and align the entire image features roughly, which may result in model collapse~\cite{tang2020discriminative} and negative transfer~\cite{li2021semantic}.
To alleviate these problems, inspired by the finding that the implicit discriminative ability of the original classifier~\cite{chen2022reusing}, and the localization ability of convolutional neural networks shown in~\cite{zhou2016learning}, we propose \underline{C}rucial \underline{S}emantic \underline{C}lassifier-based \underline{A}dversarial \underline{L}earning for UDA.
It is noted that, in terms of network architecture, we have no additional domain discriminator design. 
The feature extractor $G$ extracts the semantic features $\mathbf{f}$ from input samples, i.e., $\mathbf{f}^{s}_{i} = G(x^{s}_{i})$ and $\mathbf{f}^{t}_{j} = G(x^{t}_{j})$.
The corresponding predictions are calculated by the classifier $C$, i.e., $\mathbf{p}^{s}_{i} = C(\mathbf{f}^{s}_{i})$ and $\mathbf{p}^{t}_{j} = C(\mathbf{f}^{t}_{j})$.


\subsubsection{Overview}
The overview of CSCAL is depicted in Fig.~\ref{fig:fig2}.
Specifically, the objective function of CSCAL mainly consists of three parts.
Firstly, for the source domain samples $x_{i}^s$, we can calculate the cross-entropy $\mathcal{L}_{ce}$ with the labels $y_{i}^s$ to optimize classifier $C$.
Secondly, for each class, we construct the intra-domain paired samples $\{\mathbf{p}^s_{i}, \mathbf{p}^s_{i'}\}_{y^{s}_{i}=y^{s}_{i'}}$ and inter-domain paired samples $\{\mathbf{p}^s_{k}, \mathbf{p}^t_{j}\}_{y^{s}_{k} = y^{p}_{j}}$ according to their corresponding ground-truth labels $y^{s}_{k}$, and pseudo labels $y^{p}_{j}$ (i.e., the predictions of $x_j^t$), then the Paired-Level Discrepancy (PLD) $\mathcal{L}_{pld}$ can be measured and optimized to achieve crucial semantic knowledge transferring.
In our adversarial paradigm, classifier $C$ strives to maximize $\mathcal{L}_{pld}$, while feature extractor $G$ encourages minimizing it. 
Since the probabilities of the correct class are similar, the main discrepancies are mainly present in wrong predictions.
Thus, the weights of wrong predictions are encouraged to improve when classifier $C$ is training. 
Because of the adversarial mechanism, the semantic features of wrong predictions are suppressed while crucial semantic features are highlighted in the feature extractor $G$ training procedure.
In this way, we can filter the semantic features required for transferring.
Moreover, except for intra-class-wise alignment, the inter-class-wise information should not be ignored, which can help reduce ambiguous predictions and further adaptation performance.
Given this, we leverage the source and target prediction matrics to construct a Nuclear Norm-based Discrepancy (NND) as the adversarial domain critic. 
Unlike traditional works utilizing tedious alter-stage training strategies, we complete our method via the Gradient Reverse Layer (GRL), as shown in Fig.~\ref{fig:fig2}.

\begin{figure}
    \centering
    \includegraphics[width=1.0\linewidth]{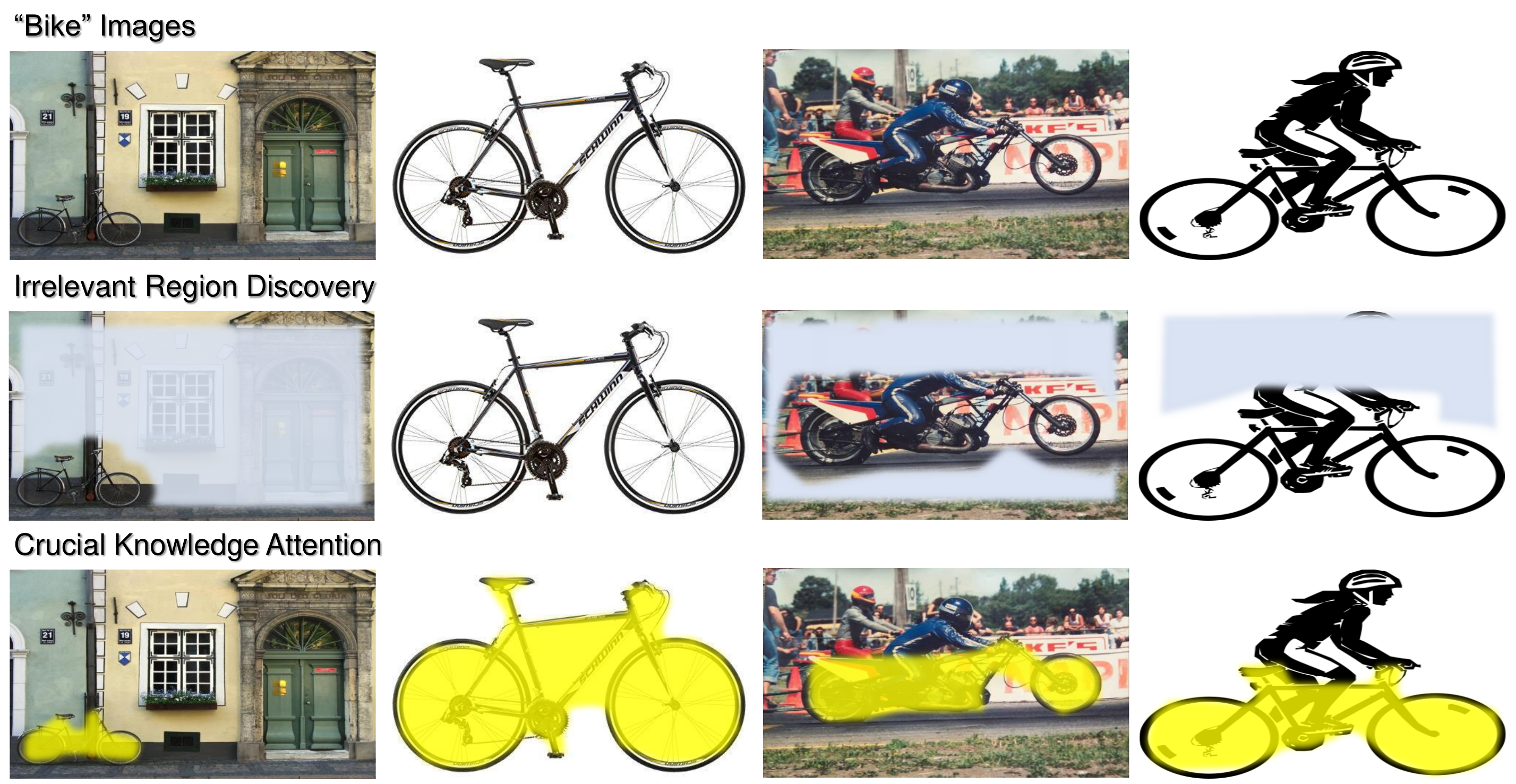}
    \caption{Illustration of Crucial Semantic Knowledge Attention procedure.}
    \label{fig:fig3}
\end{figure}

\subsection{Crucial Semantic Knowledge Attention}
Existing adversarial-based methods~\cite{long2018conditional, tzeng2017adversarial, long2017deep} directly adapt entire semantic features from source to target, neglecting the negative transfer caused by tedious irrelevant information. 
Motivated by the attention mechanism of DNNs~\cite{zhou2016learning}, we achieve crucial semantic attention and transfer via an intra-class-wise adversarial optimizing paradigm in which the original classifier $C$ plays the role of the discriminator.
Specifically, there are two steps to achieving our goal, as shown in Fig.~\ref{fig:fig3}, and our ideas are explained as follows.

\subsubsection{Irrelevant Region Discovery}
As shown in Fig.~\ref{fig:fig2}, the paired samples can be formed according to their ground-truth labels $y_i^{s}$ and pseudo labels $y_j^p$.
For each category, the PLD $\mathcal{L}_{pld}$ can be decomposed as intra-domain part $\mathcal{L}_{pld}^{intra}$ and inter-domain part $\mathcal{L}_{pld}^{inter}$:
\begin{equation}
\begin{aligned}
    \mathcal{L}_{pld}  &= \mathcal{L}_{pld}^{intra} + \mathcal{L}_{pld}^{inter} \\
    &= \mathbb{E}[\mathrm{JS}(\mathbf{p}^{s}_{i},\mathbf{p}^{s}_{i'})]  + \mathbb{E}[\mathrm{JS}(\mathbf{p}^{s}_{k},\mathbf{p}^{t}_{j})].
\end{aligned}
\end{equation}
Here, we use Jensen-Shannon divergence $\mathrm{JS}$~\cite{fuglede2004jensen} due to its symmetry, which Kullback-Leibler divergence~\cite{dong2022federated} does not have, to measure the discrepancy across samples in the same pair, where $y_{i}^{s}=y_{i'}^s$ and $y_{k}^s=y_{j}^{p}$.

For each class paired samples, as shown in Fig.~\ref{fig:fig2}, their predictive scores are similar in that class, while the main discrepancy is across other wrong predictions.
Inspired by this observation, we encourage classifier $C$ to maximize the $\mathcal{L}_{pld}$:
\begin{equation}
    \max_{C}\mathcal{L}_{pld} = \mathbb{E}[\mathrm{JS}(\mathbf{p}^{s}_{i},\mathbf{p}^{s}_{i'})]  + \mathbb{E}[\mathrm{JS}(\mathbf{p}^{s}_{k},\mathbf{p}^{t}_{j})].
\end{equation}
As a result, the weights of wrong predictions are improved, which drives the corresponding irrelevant semantic features activated.
In this way, we have discovered the irrelevant region.

\subsubsection{Crucial Knowledge Attention}
Since the irrelevant region has been discovered in the previous subsection, we can achieve crucial knowledge attention via irrelevant semantic features suppressed.
From this perspective, we optimize feature extractor $G$ to minimize $\mathcal{L}_{lpd}$:
\begin{equation}
    \min_{G}\mathcal{L}_{pld} = \mathbb{E}[\mathrm{JS}(\mathbf{p}^{s}_{i},\mathbf{p}^{s}_{i'})]  + \mathbb{E}[\mathrm{JS}(\mathbf{p}^{s}_{k},\mathbf{p}^{t}_{j})].
\end{equation}
In previous classifier training, the irrelevant semantic features were activated. 
Thus the feature extractor is encouraged to mitigate such discrepancies and strengthen their similar part, which means crucial semantic knowledge is highlighted.
In our adversarial manner, the model is optimized via the:
\begin{equation}
    \min_{G}\max_{C}\mathcal{L}_{pld}.
\end{equation}

Therefore, the model adaptation can be more effective via our proposed crucial semantic knowledge attention mechanism.
In more detail, the domain shift is mitigated by aligning inter-domain PLD, and crucial semantic knowledge attention is mainly achieved by aligning intra-domain PLD.

\subsection{Nuclear Norm-based Discrepancy}
Despite aligning PLD via an adversarial learning manner can achieve effective adaptation. 
However, only considering intra-class-wise while neglecting inter-class-wise discrepancy may generate inaccurate results~\cite{chen2022reusing}. 
Thus, to further adaptation performance, we add inter-class-wise alignment relying on the prediction matrics from the classifier as domain critic.

Review previous adversarial-based methods~\cite{ganin2015unsupervised, long2017deep, tzeng2017adversarial,long2018conditional} that leverage a well-designed domain discriminator to drive the generator to mitigate domain shift by optimizing adversarial domain loss. WGAN~\cite{arjovsky2017wasserstein} was further introduced to learn a critic function $h$ by optimizing Wasserstein distance between source representation distributions $\mathbb{P}^{s}$ and target representation distributions $\mathbb{P}^{t}$:
\begin{equation}
    W(\mathbb{P}^{s},\mathbb{P}^{t}) = \sup_{\Vert h\Vert_{L}\leq 1}\mathbb{E}_{f^s \sim \mathbb{P}^s}[h(f^s)]-\mathbb{E}_{f^t \sim \mathbb{P}^t}[h(f^t)], 
    \label{eq5}
\end{equation}
where $f^s$ and $f^t$ denote the representations extracted from the source and target domain, respectively, and $\Vert\cdot\Vert$is the Lipschitz semi-norm. Inspired by WGAN~\cite{arjovsky2017wasserstein}, we leverage classifier $C$ to play the critic function role and analyze the possibility.
\subsubsection{Rethinking the classifier output}
Since absent supervised information and domain shift, there are more wrong predictions on the target domain while more accurate predictions on the source domain.
There are reasons to believe that the error rate $\mathcal{E}_t$ on target is higher than the error rate $\mathcal{E}_s$ on the source, while the accurate rate $\mathcal{A}_t$ on the target is lower than the accurate rate $\mathcal{A}_s$ on the source, and they satisfy the inequality:
\begin{equation}
    \Vert \mathcal{A}_{s} - \mathcal{E}_{s} \Vert > \Vert \mathcal{A}_{t} - \mathcal{E}_{t} \Vert.
\end{equation}
Hence, the discrepancy between the source and target domain can be revealed via the difference between accurate rate $\mathcal{A}$ and error rate $\mathcal{E}$.
For the mini-batch input $x$, the prediction matrix $\mathcal{M} \in \mathbb{R}^{b\times k}$ can be calculated via classifier $C$, where $b$ is the batch size, and $k$ is the number of categories. 
The probability $\mathcal{M}_{i,j}$ tells the relationship between the $i$-th sample and $j$-class, and the class correlation matrix $\mathcal{S}\in\mathbb{R}^{k\times k}$ can be denoted as:
\begin{equation}
    \mathcal{S} = \mathcal{M}^\top \mathcal{M},
\end{equation}
where $\sum_{i,j}^{k} \mathcal{S}_{i,j} = b$ and $\sum_{j}^{k}\mathcal{S}_{i,j}=1$.
Actually, the class correlation matrix $\mathcal{S}$ is the coarse estimation of the class confusion matrix~\cite{jin2020minimum}.
Combined with the previous analysis, the $\Vert \mathcal{A} - \mathcal{E} \Vert$ can be revealed by $\Vert tr(\mathcal{S}) - \sum_{i\neq j}^k\mathcal{S}_{i,j} \Vert = 2\Vert tr(\mathcal{S}) \Vert - b$.
Moreover, $tr(\mathcal{S}) = \Vert \mathcal{M} \Vert_{F}$, where $\Vert \cdot \Vert_{F}$ denote Frobenius norm operation, and $tr(\mathcal{S})$ is the trace of $\mathcal{S}$.
Since $\mathcal{M}$ is the output of classifier $C$, removing the constant the $\Vert C \Vert_{F}$ can be directly regarded as the critic function.
\subsubsection{Domain Discrepancy Measure} As we analyzed before, the $\Vert C \Vert_{F}$ possesses the critic capability like function $h$ in Eq.~\ref{eq5}.
Thus, the domain discrepancy can be denoted as follows:
\begin{equation}
    \begin{aligned}
    W_{F}(\mathbb{P}^{s},\mathbb{P}^{t}) =  \sup_{\Vert \Vert C \Vert _{F}\Vert_{L}\leq 1}\mathbb{E}_{f^s \sim \mathbb{P}^s}[\Vert C(f^s)\Vert_{F}] \\ -\mathbb{E}_{f^t \sim \mathbb{P}^t}[\Vert C(f^t)\Vert_{F}].
    \label{eq8}
    \end{aligned}
\end{equation}
However, we could not directly calculate the rank of $\mathcal{M}$ since it is an NP-hard non-convex problem.
Inspired by the previous demonstrated works~\cite{cui2020towards,recht2010guaranteed} that when $\Vert \mathcal{M} \Vert_{F}\leq 1$, 
the nuclear-norm $\Vert \mathcal{M} \Vert_{n}$ is the convex envelope of the rank of $\mathcal{M}$.
Maximizing $\Vert \mathcal{M} \Vert_{n}$ means improving the prediction diversity when $\Vert \mathcal{M} \Vert_{F}$ is near $\sqrt{b}$.
Thus, in Eq.~\ref{eq8}, we replace $\Vert C \Vert_{F}$ with $\Vert C \Vert_{n}$ and rewrite the domain discrepancy as:
\begin{equation}
\begin{aligned}
    W_{n}(\mathbb{P}^{s},\mathbb{P}^{t}) = \sup_{\Vert \Vert C \Vert_{n} \Vert_{L}\leq 1}\mathbb{E}_{f^s\sim \mathbb{P}^s}[\Vert C(f^s) \Vert_{n}]\\
    -\mathbb{E}_{f^t\sim \mathbb{P}^t}[\Vert C(f^t)\Vert_{n}].
\end{aligned}
\end{equation}
In application, the Nuclear Norm-based Discrepancy (NND) $\mathcal{L}_{nnd}$ is maximized as the estimation of $W_{n}(\mathbb{P}^s, \mathbb{P}^t)$, and the definition of $\mathcal{L}_{nnd}$ is:
\begin{equation}
    \mathcal{L}_{nnd} = \mathbb{E}_{x_{i}^s\in \mathcal{D}_s}[C(D(x_i^s))] - \mathbb{E}_{x_{j}^t\in\mathcal{D}_t}[C(D(x_j^t))].
\end{equation}
Then, we can plus $\mathcal{L}_{nnd}$ in adversarial training, which also considers the inter-class divergence between predictions and further domain adaptation performance.
\begin{table*}[htbp]
    \setlength{\tabcolsep}{5pt}
    \centering
    \caption{Classification accuracy (\%) on Office-Home for UDA (ResNet-50). $^\dag$ denotes that results are reproduced according to the public source code.}
    \begin{tabular}{|l|c c c c c c c c c c c c c|}
    \hline
         Method & Ar$\rightarrow$Cl & Ar$\rightarrow$Pr & Ar$\rightarrow$Rw & Cl$\rightarrow$Ar & Cl$\rightarrow$Pr & Cl$\rightarrow$Rw & Pr$\rightarrow$Ar & Pr$\rightarrow$Cl & Pr$\rightarrow$Rw&
         Rw$\rightarrow$Ar & Rw$\rightarrow$Cl & Rw$\rightarrow$Pr & Avg\\
         \hline
         \hline
         ResNet-50~\cite{he2016deep}& 34.9 & 50.0 & 58.0 & 37.4 & 41.9 & 46.2 & 38.5 & 31.2 & 60.4 & 53.9 & 41.2 & 59.9 & 46.1\\ 
         WDGRL~\cite{shen2018wasserstein} & 44.1 & 63.8 & 74.0 & 47.3 & 57.1 & 61.7 & 51.8 & 39.1 & 72.1 & 64.9 & 45.9 & 76.5 & 58.2 \\
         MCD~\cite{saito2018maximum} & 48.9 & 68.3 & 74.6 & 61.3 & 67.6 & 68.8 & 57.0 & 47.1 & 75.1 & 69.1 & 52.2 & 79.6 & 64.1 \\
         BSP~\cite{chen2019transferability} & 52.0 & 68.6 & 76.1 & 58.0 & 70.3 & 70.2 & 58.6 & 50.2 & 77.6 & 72.2 & 59.3 & 81.9 & 66.3 \\
         BNM~\cite{cui2020towards} & 52.3 & 73.9 & 80.0 & 63.3 & 72.9 & \textcolor{blue}{74.9} & 61.7 & 49.5 & 79.7 & 70.5 & 53.6 & 82.2 & 67.9 \\
         ETD~\cite{li2020enhanced} & 51.3 & 71.9 & \textcolor{red}{\textbf{85.7}} & 57.6 & 69.2 & 73.7 & 57.8 & 51.2 & 79.3 & 70.2 & 57.5 & 82.1 & 67.3\\ 
         SymNets~\cite{zhang2019domain} & 47.7 & 72.9 & 78.5 & 64.2 & 71.3 & 74.2 & 64.2 & 48.8 & 79.5 & 74.5 & 52.6 & 82.7 & 67.6\\      
         TSA~\cite{li2021transferable} & 53.6 & 75.1 & 78.3 & 64.4 & 73.7 & 72.5 & 62.3 & 49.4 & 77.5 & 72.2 & 58.8 & 82.1 & 68.3 \\
         SCDA$^\dag$~\cite{li2021semantic} & 55.6 & 77.0 & 79.8 & 65.6 & 74.1 & 74.7 & 64.2 & 54.1 & 79.7 & 74.0 & 59.7 & 83.6 & 70.2\\
         \hline
         CSCAL & \textcolor{blue}{56.8} & \textcolor{blue}{77.4} & 80.4 & \textcolor{blue}{66.4} & \textcolor{blue}{75.6} & 74.8 & 65.5 & 53.2 & 79.9 & \textcolor{blue}{74.3} & 59.3 & \textcolor{blue}{84.3} & \textcolor{blue}{70.7}\\
         \hline
         DANN~\cite{ganin2015unsupervised} & 45.6 & 59.3 & 70.1 & 47.0 & 58.5 & 60.9 & 46.1 & 43.7 & 68.5 & 63.2 & 51.8 & 76.8 & 57.6 \\
         DANN+CSCAL & 54.2 & 66.5 & 76.4 & 60.8 & 69.7 & 70.5 & 61.6 & \textcolor{blue}{55.5} & \textcolor{blue}{80.3} & 74.0 & \textcolor{red}{\textbf{61.8}} & 82.7 & 67.8\\
         \hline
         JANN~\cite{long2017deep} & 45.9 & 61.2 & 68.9 & 50.4 & 59.7 & 61.0 & 45.8 & 43.4 & 70.3 & 63.9 & 52.4 & 76.8 & 58.3\\
         JANN+CSCAL & 49.3 & 70.6 & 76.4 & 58.6 & 66.1 & 68.7 & 62.6 & 49.2 & 77.4 & 71.4 & 53.7 & 80.9 & 65.4 \\
         \hline
         MCC~\cite{jin2020minimum} & 55.1 & 75.2 & 79.5 & 63.3 & 73.2 & 75.8 & \textcolor{blue}{66.1} & 52.1 & 76.9 & 73.8 & 58.4 & 83.6 & 69.4 \\
         MCC+CSCAL & \textcolor{red}{\textbf{56.9}} & \textcolor{red}{\textbf{80.0}} & \textcolor{blue}{82.9} & \textcolor{red}{\textbf{67.1}} & \textcolor{red}{\textbf{77.4}} & \textcolor{red}{\textbf{77.8}} & \textcolor{red}{\textbf{67.1}} & \textcolor{red}{\textbf{55.7}} & \textcolor{red}{\textbf{81.8}} & \textcolor{red}{74.8} & \textcolor{blue}{61.7} & \textcolor{red}{\textbf{85.7}} & \textcolor{red}{\textbf{72.4}} \\
         \hline
    \end{tabular}
    \label{tab:tab1}
\end{table*}
\begin{table*}[htbp]
\vspace{-0.2cm} 
\setlength{\tabcolsep}{1.3pt}
    \caption{
    Classification accuracy (\%) on DomainNet for UDA (ResNet-101). In each sub-table, the column-wise domains are selected as the source domain and the row-wise domains are selected as the target domain}
    \begin{tabular}{|c|c c c c c c c||c|c c c c c c c||c|c c c c c c c|}
        \hline
         \textbf{ResNet-101}\cite{he2016deep}& clp & inf & pnt & qdr & rel & skt & Avg. & \textbf{ADDA}\cite{tzeng2017adversarial}& clp & inf & pnt & qdr & rel & skt & Avg. & \textbf{MCD}\cite{saito2018maximum}& clp & inf & pnt & qdr & rel & skt & Avg.  \\
        \hline
        \hline
        clp & - & 19.3 & 37.5 & 11.1 & 52.2 & 41.0 & 32.2 & clp & - & 11.2 & 24.1 & 3.2 & 41.9 & 30.7 & 22.2 & clp & - & 14.2 & 26.1 & 1.6 & 45.0 & 33.8 & 24.1 \\
        inf & 30.2 & - & 31.2 & 3.6 & 44.0 & 27.9 & 27.4 & inf & 19.1 & - & 16.4 & 3.2 & 26.9 & 14.6 & 16.0 & inf & 23.6 & - & 21.2 & 1.5 & 36.7 & 18.0 & 20.2\\
        pnt & 39.6 & 18.7 & - & 4.9 & 54.5 & 36.3 & 30.8 & pnt & 31.2 & 9.5 & - & 8.4 & 39.1 & 25.4 & 22.7 & pnt & 34.4 & 14.8 & - & 1.9 & 50.5 & 28.4 & 26.0\\
        qdr & 7.0 & 0.9 & 1.4 & - & 4.1 & 8.3 & 4.3 & qdr & 15.7 & 2.6 & 5.4 & - & 9.9 & 11.9 & 9.1 & qdr & 15.0 & 3.0 & 7.0 & - & 11.5 & 10.2 & 9.3\\
        rel & 48.4 & 22.2 & 49.4 & 6.4 & - & 38.8 & 33.0 & rel & 39.5 & 14.5 & 29.1 & 12.1 & - & 25.7 & 24.2 & rel & 42.6 & 19.6 & 42.6 & 2.2 & - & 29.3 & 27.2 \\
        skt & 46.9 & 15.4 & 37.0 & 10.9 & 47.0 & - & 31.4 & skt & 35.3 & 8.9 & 25.2 & 14.9 & 37.6 & - & 25.4 & skt & 41.2 & 13.7 & 27.6 & 3.8 & 34.8 & - & 24.2\\
        Avg. & 34.4 & 15.3 & 31.3 & 7.4 & 40.4 & 30.5 & \textbf{26.6} & Avg. & 28.2 & 9.3 & 20.1 & 8.4 & 31.1 & 21.7 & \textbf{19.8} & Avg. & 31.4 & 13.1 & 24.9 & 2.2 & 35.7 & 23.9 & \textbf{21.9}\\
        \hline
        \hline        \textbf{SCDA}\cite{li2021semantic} & clp & inf & pnt & qdr & rel & skt & Avg. & \textbf{DANN}\cite{ganin2015unsupervised} & clp & inf & pnt & qdr & rel & skt & Avg. & \textbf{CDAN}\cite{long2018conditional} & clp & inf & pnt & qdr & rel & skt & Avg.\\
        \hline
        \hline
        clp & - & 18.6 & 34.8 & 9.5 & 50.8 & 41.4 & 32.4 & clp &- & 14.2 & 26.1 & 1.6 & 45.0 & 33.8 & 24.1 & clp & - & 20.4 & 36.6 & 9.0 & 50.7 & 42.3 & 31.8\\
        inf & 29.6 & - & 34.0 & 1.4 & 46.3 & 25.4 & 27.3 & inf & 31.8 & - & 30.2 & 3.8 & 44.8 & 25.7 & 27.3 & inf &27.5 & - & 25.7 & 1.8 & 34.7 & 20.1 & 22.0  \\
        pnt & 44.1 & 19.0 & - & 2.6 & 56.2 & 42.0 & 32.8 & pnt & 39.6 & 15.1 & - & 5.5 & 54.6 & 35.1 & 30.0 & pnt & 42.6 & 20.0 & - & 2.5 & 55.6 & 38.5 & 31.8  \\
        qdr & 30.0 & 4.9 & 15.0 & - & 25.4 & 19.8 & 19.0 & qdr & 11.8 & 2.0 & 4.4 & - & 9.8 & 8.4 & 7.3 & qdr & 21.0 & 4.5 & 8.1 & - & 14.3 & 15.7 & 12.7 \\
        rel & 54.0 & 22.5 & 51.9 & 2.3 & - & 42.5 & 34.6 & rel & 47.5 & 17.9 & 47.0 & 6.3 & - & 37.3 & 31.2 & rel & 51.9 & 23.3 & 50.4 & 5.4 & - & 41.4 & 34.5 \\
        skt & 55.6 & 18.5 & 44.7 & 6.4 & 53.2 & - & 35.7& skt & 47.9 & 13.9 & 34.5 & 10.4 & 46.8 & - & 30.7 & skt & 50.8 & 20.3 & 43.0 & 2.9 & 50.8 & - & 33.6\\
        Avg.& 42.6 & 16.7 & 37.0 & 3.6 & 47.2 & 34.8 & \textbf{30.3} & Avg. & 35.7 & 12.9 & 30.2 & 7.1 & 41.4 & 29.6 & \textbf{26.1} & Avg. & 38.8 & 17.7 & 32.8 & 4.3 & 41.2 & 31.6 & \textbf{27.7} \\
        \hline
        \hline
        \textbf{CSCAL} & clp & inf & pnt & qdr & rel & skt & Avg. & \textbf{DANN+CSCAL} & clp & inf & pnt & qdr & rel & skt & Avg. & \textbf{CDAN+CSCAL} & clp & inf & pnt & qdr & rel & skt & Avg.\\
        \hline
        clp & - & 18.6 & 39.9 & 5.1 & 55.9 & 44.0 & 32.7 & clp & - & 21.1 & 39.4 & 13.6 & 56.1 & 45.9 & 35.2 & clp & - & 20.3 & 40.2 & 8.0 & 55.5 & 44.9 & 33.8\\
        inf & 31.0 & - & 34.9 & 1.4 & 46.1 & 26.9 & 28.1 & inf & 34.0 & - & 31.8 & 6.8 & 47.0 & 27.4 & 29.4 & inf & 32.1 & - & 32.5 & 4.6 & 46.7 & 27.5 & 28.7 \\
        pnt & 44.3 & 18.8 & - & 1.1 & 56.5 & 42.7 & 32.5 & pnt & 42.4 & 19.7 & - & 7.4 & 56.6 & 39.5 & 33.1 & pnt & 44.4 & 19.5 & - & 5.0 & 57.3 & 39.8 & 33.2 \\
        qdr & 30.3 & 4.8 & 15.7 & - & 24.4 & 20.2 & 19.1 & qdr & 22.5 & 4.2 & 9.1 & - & 14.9 & 15.7 & 13.3 & qdr & 22.8 & 3.5 & 8.9 & - & 16.3 & 16.8 & 13.7 \\
        rel & 54.9 & 22.4 & 52.0 & 3.7 & - & 42.5 & 35.1 & rel & 52.2 & 22.7 & 51.8 & 6.4 & - & 40.7 & 34.8 & rel & 54.8 & 24.1 & 54.2 & 5.0 & - & 41.9 & 36.0 
        \\
        skt & 55.5 & 18.8 & 45.6 & 10.4 & 53.9 & - & 36.8 & skt & 56.7 & 20.9 & 46.1 & 15.1& 54.7 & - & 38.7 & skt & 57.8 & 21.4 & 46.8 & 12.3 & 55.4 & - & 38.7\\
        Avg. & 43.2 & 16.7 & 37.6 & 4.3 & 47.4 & 35.3 &\textbf{30.7} & Avg. & 41.6 & 17.7 & 35.6 & 9.9 & 45.9 & 33.8 &\textbf{30.8} & Avg. & 42.4 & 17.8 & 36.5 & 7.0 & 46.2 & 34.2 & \textbf{30.7}\\
        \hline
    \end{tabular}
    \label{tab:tab2}
\end{table*}
\begin{figure*}[t!]
    \centering
    \vspace{-0.3cm} 
    \subfigcapskip=-8pt
    \subfigure[ResNet-50]{  \includegraphics[width=0.21\linewidth]{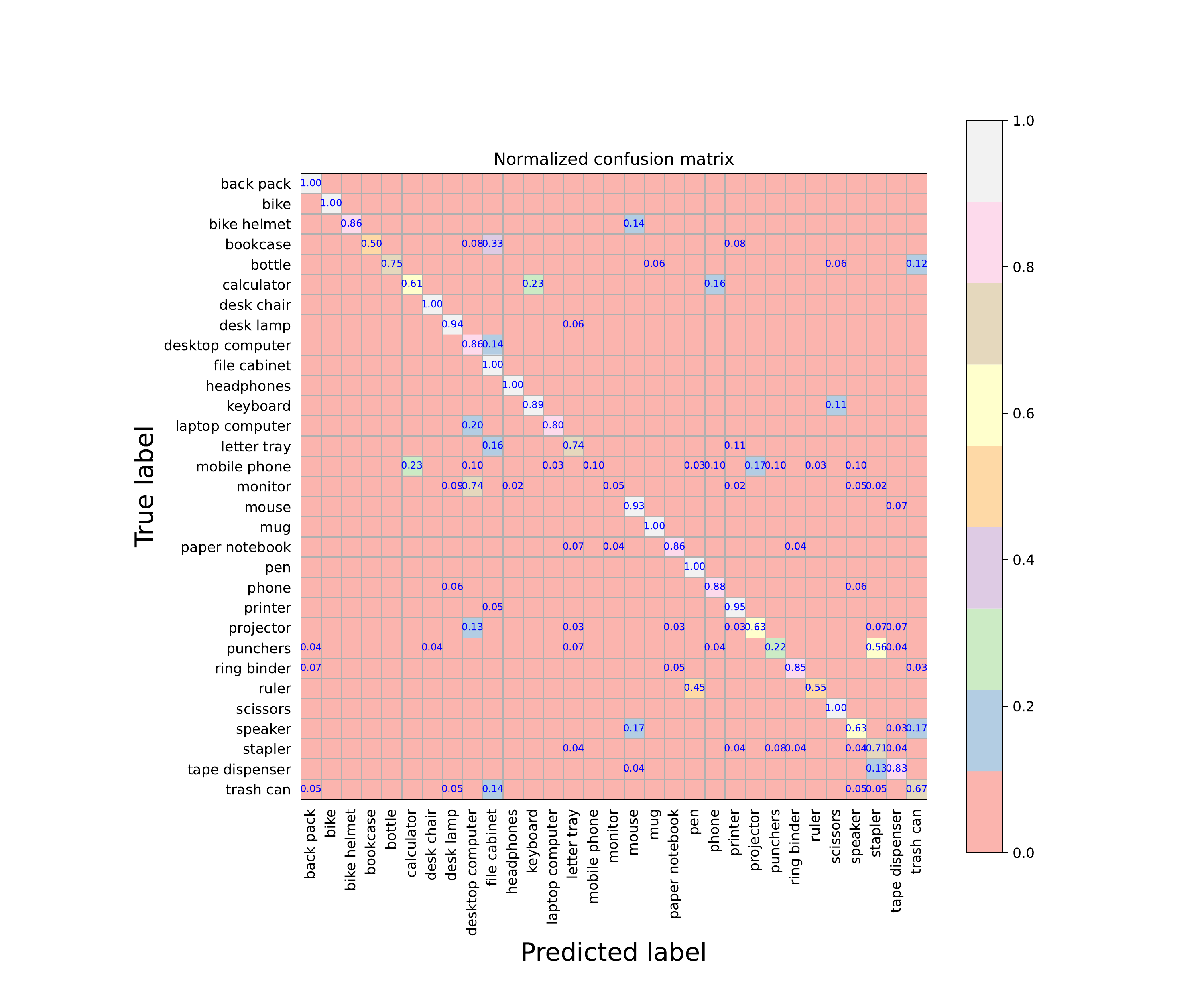}}
    \hspace{-12mm}
    \subfigure[DANN]{\includegraphics[width=0.21\linewidth]{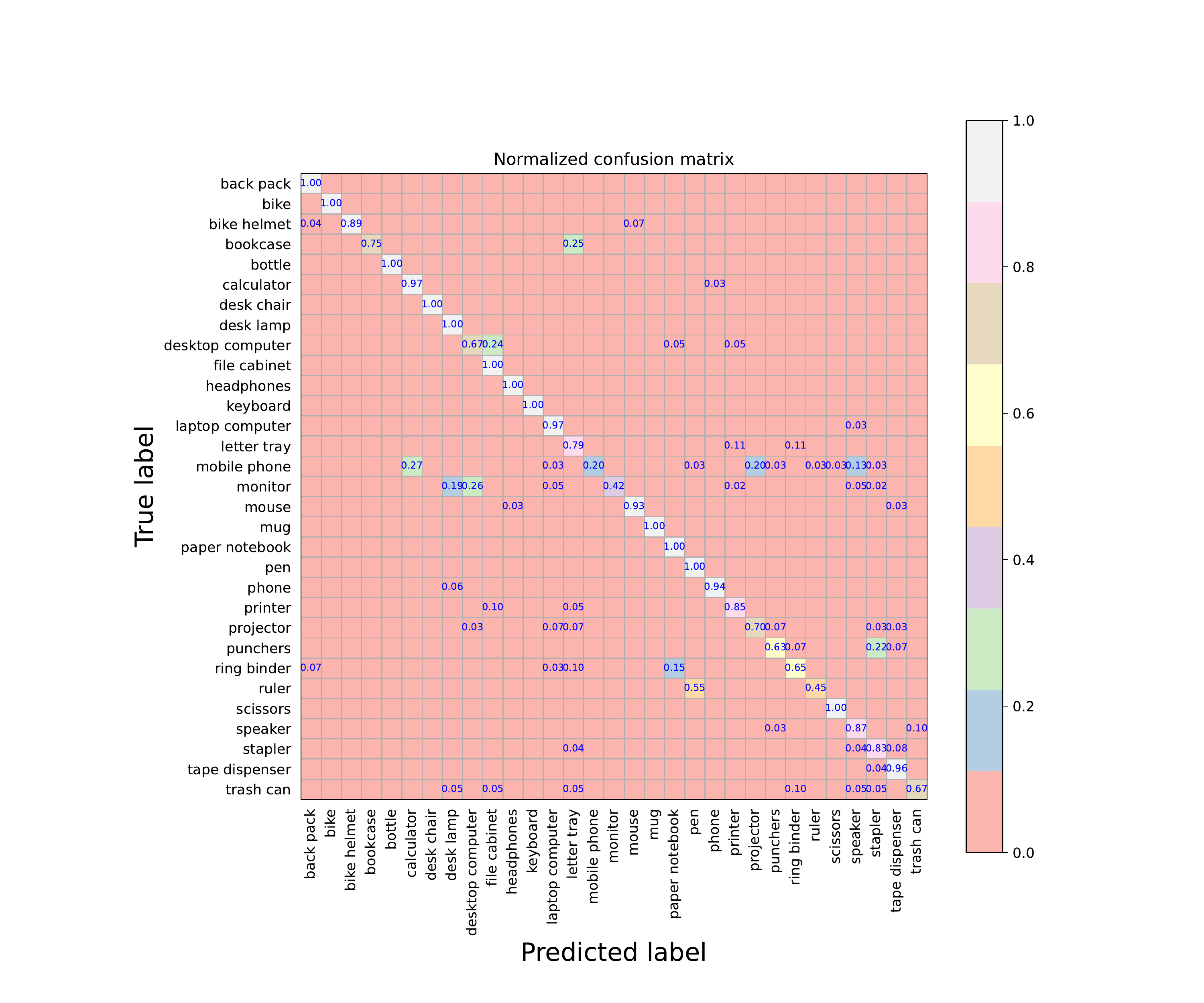}}
    \hspace{-12mm}
    \subfigure[JANN]{\includegraphics[width=0.21\linewidth]{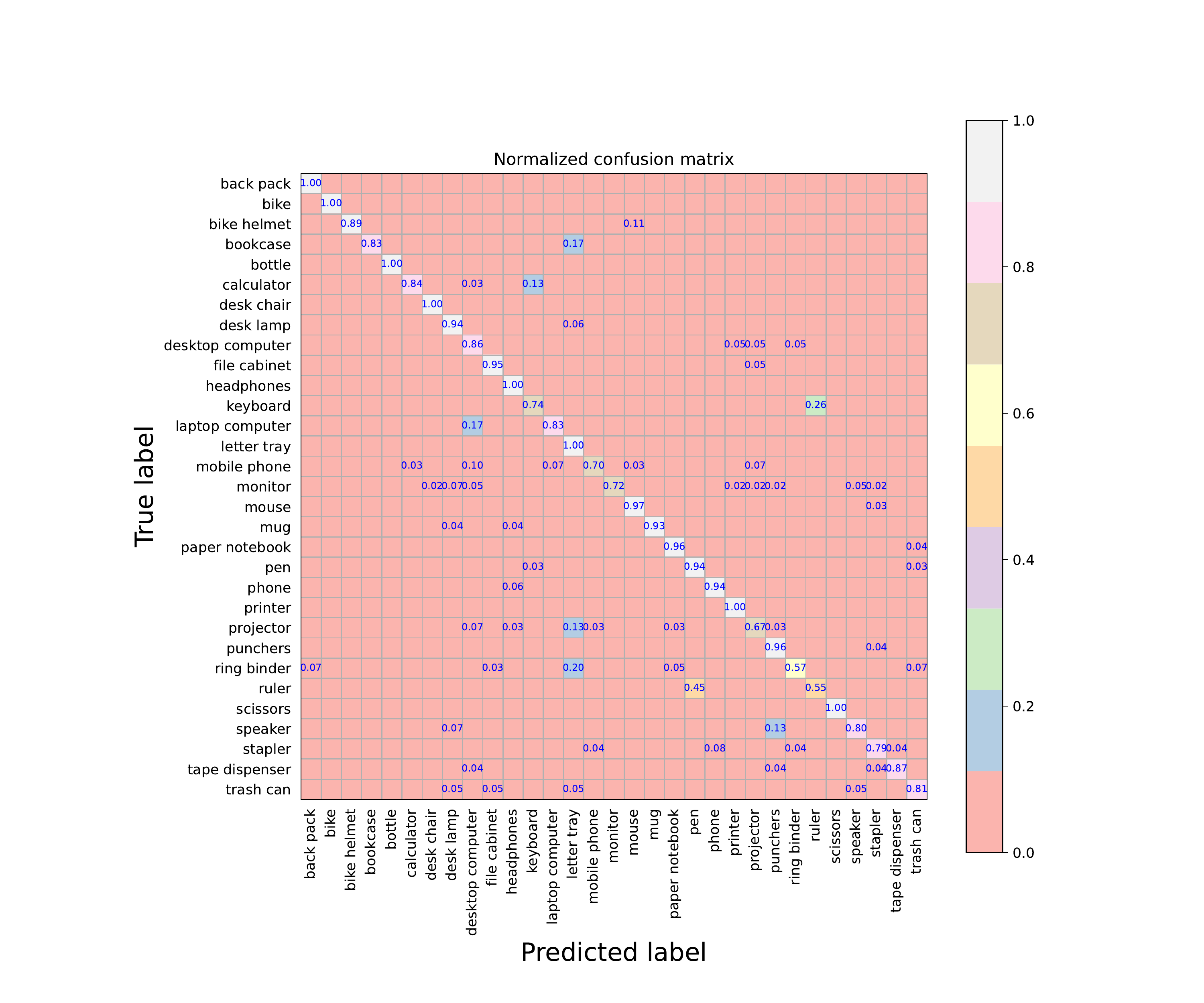}}
    \hspace{-12mm}
    \subfigure[CSCAL]{\includegraphics[width=0.21\linewidth]{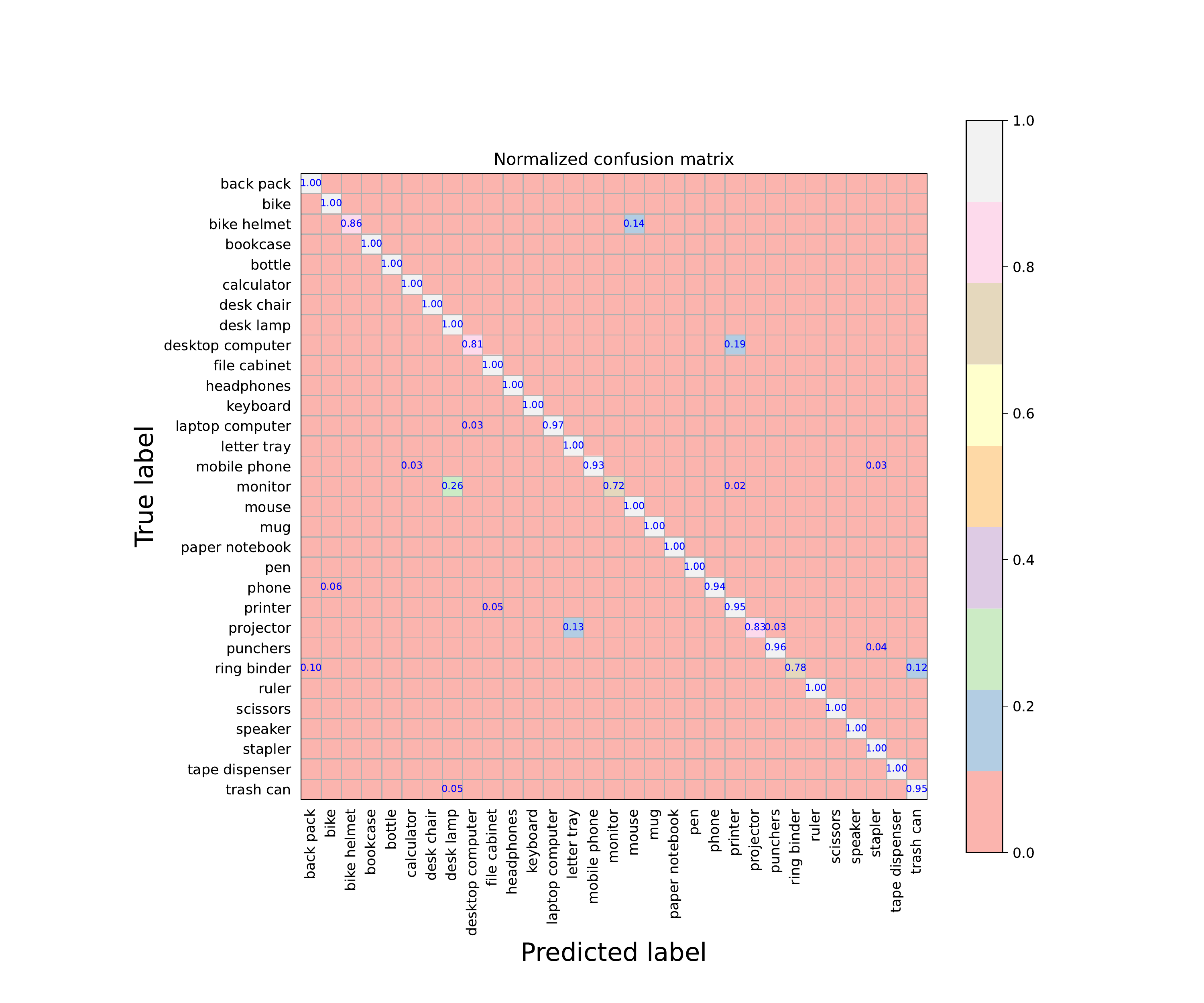}}
    \hspace{-12mm}
    \subfigure[DANN+CSCAL]{\includegraphics[width=0.21\linewidth]{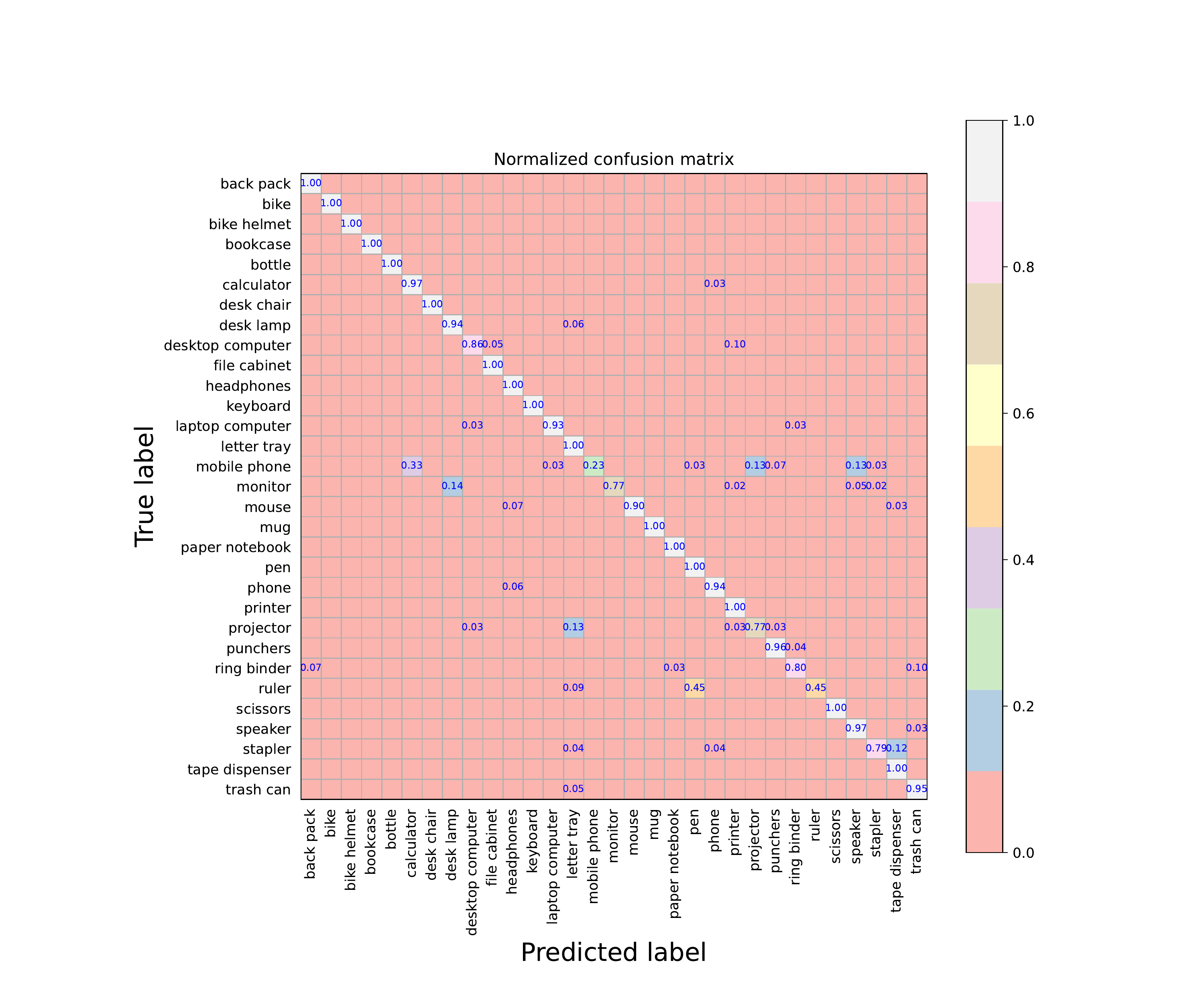}}
    \hspace{-12mm}
    \subfigure[JANN+CSCAL]{\includegraphics[width=0.21\linewidth]{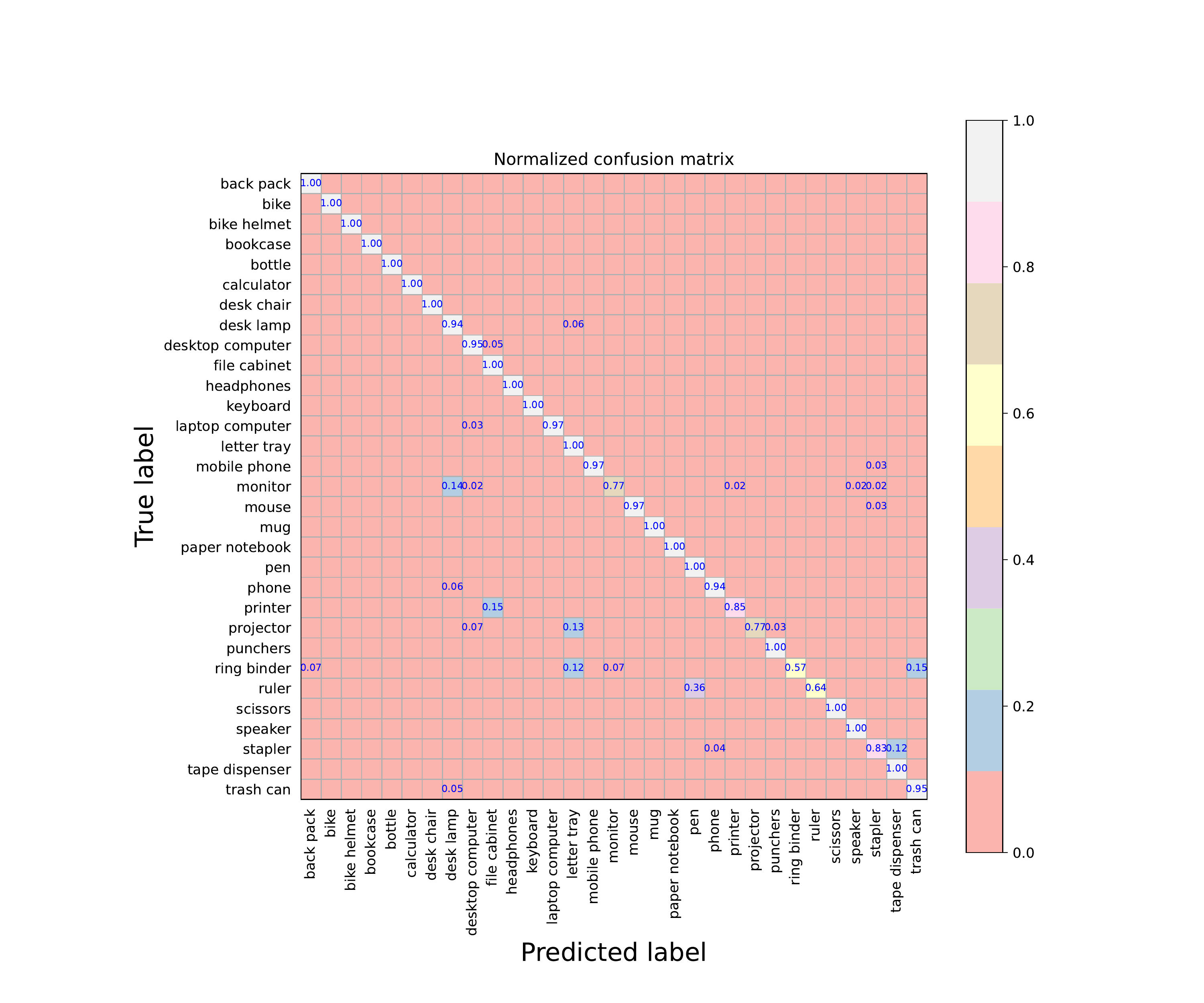}}
    \caption{The target domain confusion matrics of different methods on task A$\rightarrow$W of Office-31.}
    \label{fig:fig4}
\end{figure*}

\subsection{Overall Optimization}
To sum up, the overall objective function $\mathcal{L}_{obj}$ of CSCAL is defined as follows:
\begin{equation}
    \mathcal{L}_{obj} = \mathcal{L}_{ce} - \alpha\mathcal{L}_{pld} - \beta\mathcal{L}_{nnd}-\gamma\mathcal{L}_{mi}, 
\end{equation}
where $\alpha$, $\beta$, and $\gamma$ are hyper-parameters. $\mathcal{L}_{ce}$ is the standard cross-entropy loss on the source domain, which can be presented as:
\begin{equation}
    \mathcal{L}_{ce} = \mathbb{E}_{(x_{i}^{s}, y_{i}^s)\in\mathcal{D}_s}[-y_{i}^s\log(C(G(x_{i}^s)))].
\end{equation}
Since our inter-domain paired samples construction requires reliable pseudo labels for target data, we introduce mutual information maximization $\mathcal{L}_{mi}$ on target data to enhance the quality of pseudo labels, and $\mathcal{L}_{mi}$ is:
\begin{equation}
    \mathcal{L}_{mi} = -\sum_{k=1}^{K}\Bar{\mathbf{p}}^{(k)}\log(\Bar{\mathbf{p}}^{(k)}) + \mathbb{E}\langle\mathbf{p}^{t}_{j}, \log\mathbf{p}^{t}_{j}\rangle,
\end{equation}
where $\Bar{\mathbf{p}} = \frac{1}{N_t}\sum_{j=1}^{N_t}\mathbf{p}_{j}^{t}$, $\Bar{\mathbf{p}}^{(k)}$ is the $k$-th element of $\Bar{\mathbf{p}}$, and $\langle\cdot,\cdot\rangle$ is the inner product operation.
Particularly, CSCAL is optimized by an adversarial loss function as follows:
\begin{equation}
    \min_{C,G}\{{\mathcal{L}_{ce}-\max_{C}}[\alpha\mathcal{L}_{pld}+\beta\mathcal{L}_{nnd}]-\gamma\mathcal{L}_{mi}\}.
    \label{eq9}
\end{equation}
As shown in Fig.~\ref{fig:fig2}, we utilize the Gradient Reverse Layer (GRL)~\cite{ganin2016domain} to perform the adversarial alignment without tedious alter-stage training.

\section{Experiments}
In this section, we evaluate CSCAL as opposed to other SOTA methods for UDA classification on three benchmark datasets, i.e., Office-Home~\cite{venkateswara2017deep}, DomainNet~\cite{peng2019moment}, and Office-31~\cite{saenko2010adapting}.
Besides, we integrated CSCAL into popular UDA methods, including DANN~\cite{ganin2015unsupervised}, JANN~\cite{long2017deep}, MCC~\cite{jin2020minimum}, and CDAN~\cite{long2018conditional}, to demonstrate the effectiveness of CSCAL.

\subsection{Experimental setup}

\textbf{Office-Home~\cite{venkateswara2017deep}.}
As a large-scale benchmark dataset, Office-Home contains 15500 images and covers 65 categories drawn from four unlike domains, i.e., Artistic images (Ar), Clip Art (Cl), Product images (Pr), and Real-World images (Rw). 
There are 12 domain adaptation tasks, Ar$\rightarrow$Ci, ..., Rw$\rightarrow$Pr, constructed to evaluate our method.

\textbf{DomainNet~\cite{peng2019moment}.}
DomainNet is the largest and the most challenging dataset containing about 0.6 million images and over 345 categories, which is collected from six different domains: Clipart(clp), Infograph (inf), Painting(pnt), Quickdraw (qdr), Real(rel) and Sketch (skt). 
We construct 30 domain adaptation tasks: clp$\rightarrow$inf, ..., skt$\rightarrow$rel.

\textbf{Office-31~\cite{saenko2010adapting}.}
Office-31 is a classical benchmark dataset for evaluating domain adaptation methods. 
It has a total number of 4110 images of 3 distinct domains, Amazon (A), Webcam (W), and DSLR (D). Each of them has 31 categories. 
Hence, six domain adaptation tasks (i.e., A$\rightarrow$W, ...,  D$\rightarrow$W) are constructed to evaluate our method.

\textbf{Implementation details}
CSCAL is implemented in the PyTorch framework~\cite{paszke2019pytorch} running on RTX 3060.
Following the standard protocols~\cite{long2018conditional} in UDA classification, the labeled source and unlabeled target data are both used to optimize the model. After adaptation training, we compare methods on the unlabeled target data.
In order to fairly compare the results, for datasets: Office-Home and Office-31, we use ResNet-50~\cite{he2016deep} pre-trained on ImageNet~\cite{russakovsky2015imagenet} as the backbone; 
for DomainNet, we use ResNet-101~\cite{he2016deep} pre-trained on ImageNet~\cite{russakovsky2015imagenet} as the backbone.
The image size of input data is cropped to $224\times224$, and the Stochastic Gradient Descent (SGD) optimizer is used to optimize the model with momentum of $0.9$ and weight decay of $1e-3$.
The hyper-parameters are set as $\beta=0.1$ and $\gamma=0.1$. 
To alleviate the negative influence of inaccurate predictions in the early training phase, we set the $\alpha=\omega \times \alpha_0$, where $\alpha_0 = 1.0$, and $\omega$ is from $0$ to $1$ with the training process.

\subsection{Comparison Results}
\textbf{Results on Office-Home} have summarized in TABLE~\ref{tab:tab1}.
Compared with other SOTA methods, our method achieves the outperforming average result of 70.7\%. 
Besides, the DANN, JANN, and MCC dramatically improve when CSCAL is a regularizer.
Their performances improved by 10.2\%, 7.1\%, and 3.0\%,  respectively.
Notably, with the help of SACAL, the MCC got ten sub-tasks best accuracy and finally achieved the overall average SOTA performance of 72.4\%.

\textbf{Results on DomainNet} are shown in TABLE~\ref{tab:tab2}.
The results indicate that CSCAL performs better than other UDA methods, with an overall average of 30.7\%.
Additionally, with the help of SACAL, the performances of DANN and CDAN improved by 4.7\% and 3.0\%, respectively.

\textbf{Results on Office-31} are presented in TABLE~\ref{tab:tab3}.
Compared with other methods, CSCAL obtained a superior overall average accuracy of 90.0\%.
Moreover, the integrated methods’ performance has all been improved when CSCAL as a regularizer. 
Specifically, with the help of CSCAL, these methods are improved by 3.7\%, 3.3\%, and 0.1\% for DANN, JANN, and MCC, respectively.

\subsection{Analysis}
\textbf{Ablation Study.}
To analyze the effect of different components of CSCAL, we conducted four ablation experiments on Office-31 base ResNet-50, and the results are presented in TABLE~\ref{tab:tab4}.
Specifically, CSCAL(w/o $\mathcal{L}_{pld}$) denotes removing the $\mathcal{L}_{pld}$ part, while CSCAL(w/o $\mathcal{L}_{nnd}$) and CSCAL(w/o $\mathcal{L}_{mi}$) indicate similar expressions.
Without $\mathcal{L}_{mi}$, the quality of pseudo labels is affected making the model cannot construct reliable inter-domain paired samples.
CSCAL outperforms CSCAL(w/o $\mathcal{L}_{nnd}$) due to inter-class discrepancy also aligned.

\textbf{Confusion Matrix.}
We present the confusion matrics of different methods in Fig.~\ref{fig:fig4}.
It can be seen obviously that there is enormous misclassification appearing in the ResNet-50 confusion matrix, and DANN and JANN also need clarification on their predictions.
On the contrary, benefiting from our proposed adversarial paradigm, CSCAL generates more reliable predictions. 
Besides, CSCAL also helps to reduce the off-diagonal elements of DANN and JANN, which shows the advantages of our method.
\begin{figure}
    \centering
    \subfigcapskip=-8pt
    \subfigure[ResNet-50]{\includegraphics[width=0.32\linewidth]{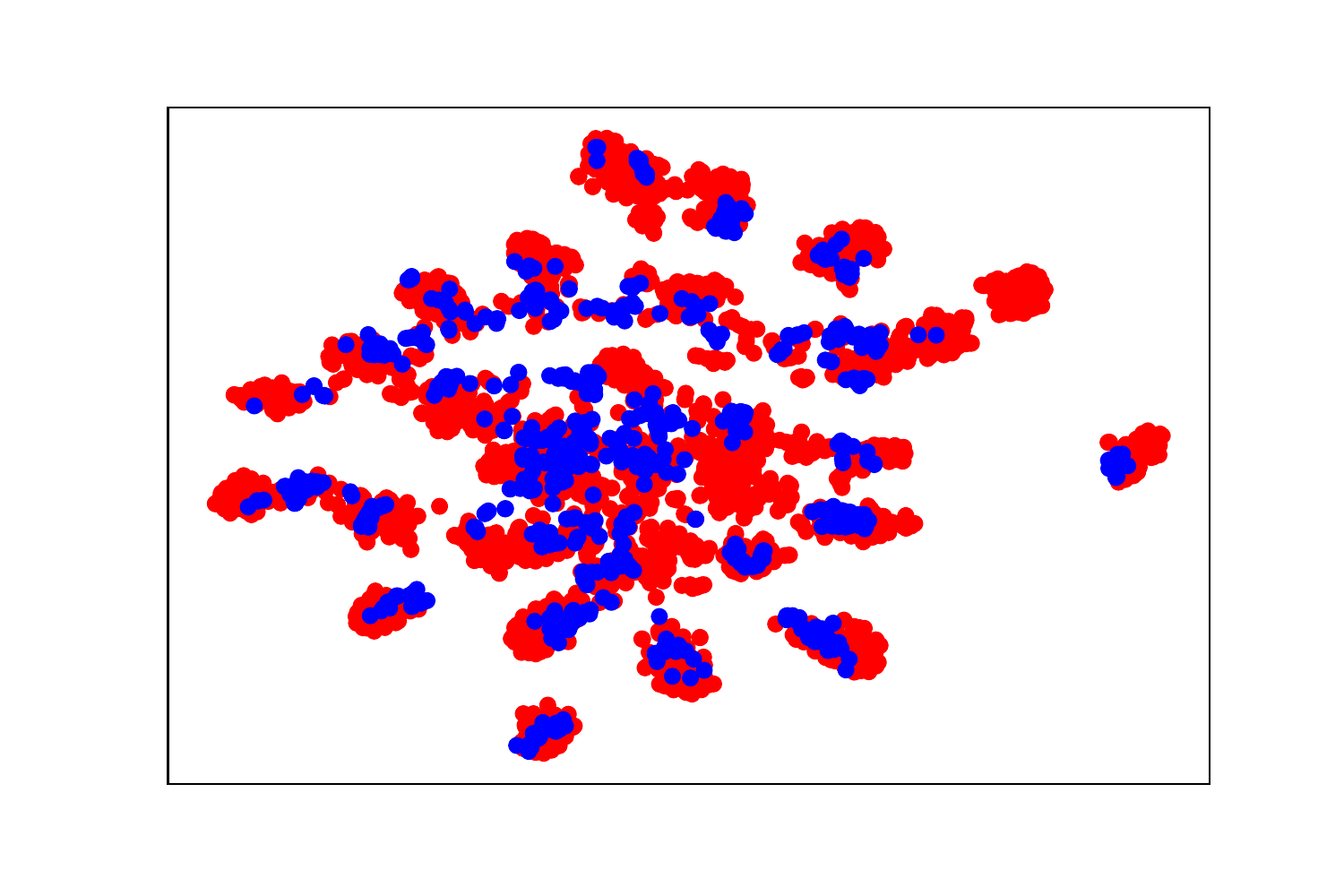}}
    \vspace{-3mm}
    \hspace{-5mm}
    \subfigure[DANN]{\includegraphics[width=0.32\linewidth]{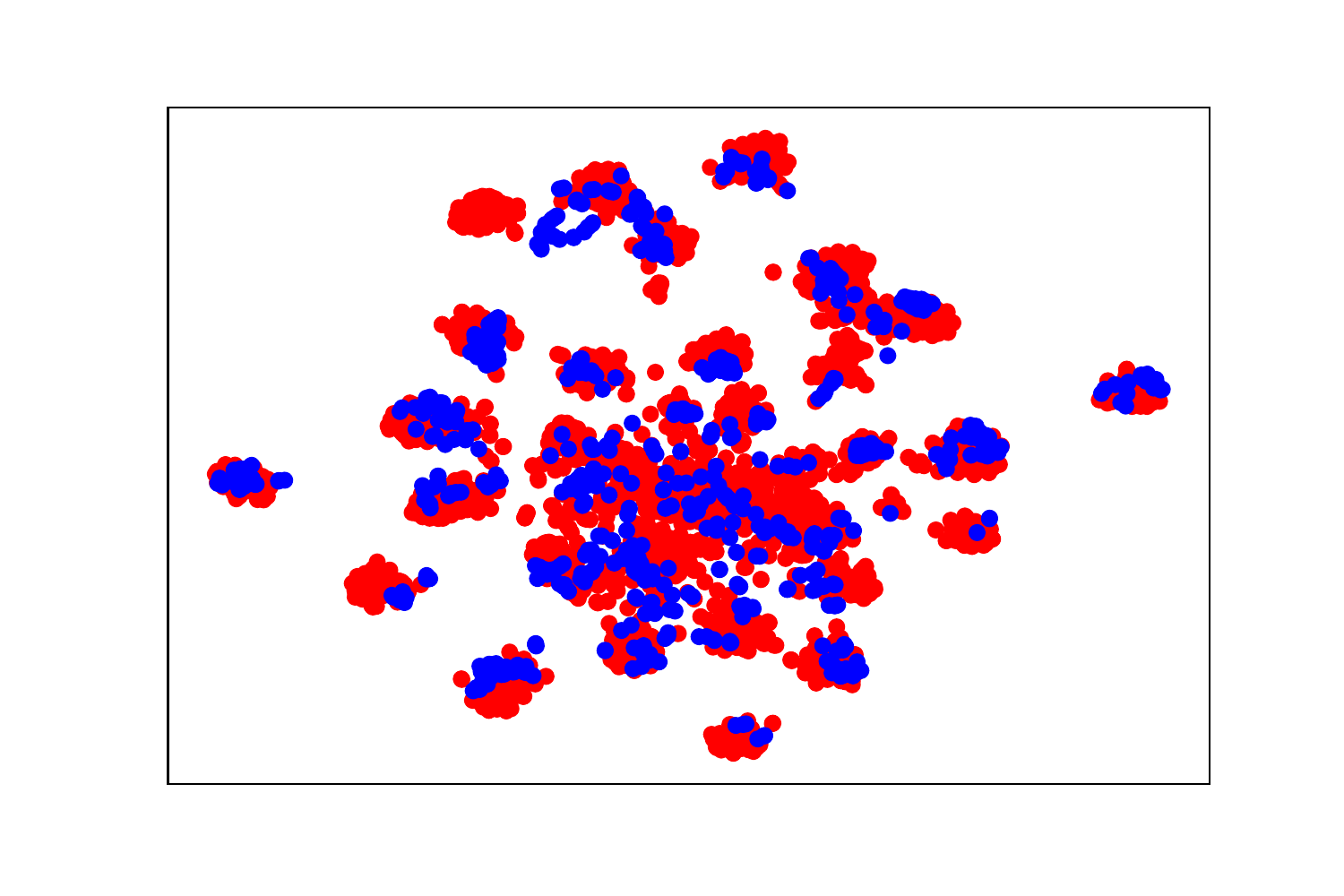}}
    \hspace{-5mm}
    \subfigure[JANN]{\includegraphics[width=0.32\linewidth]{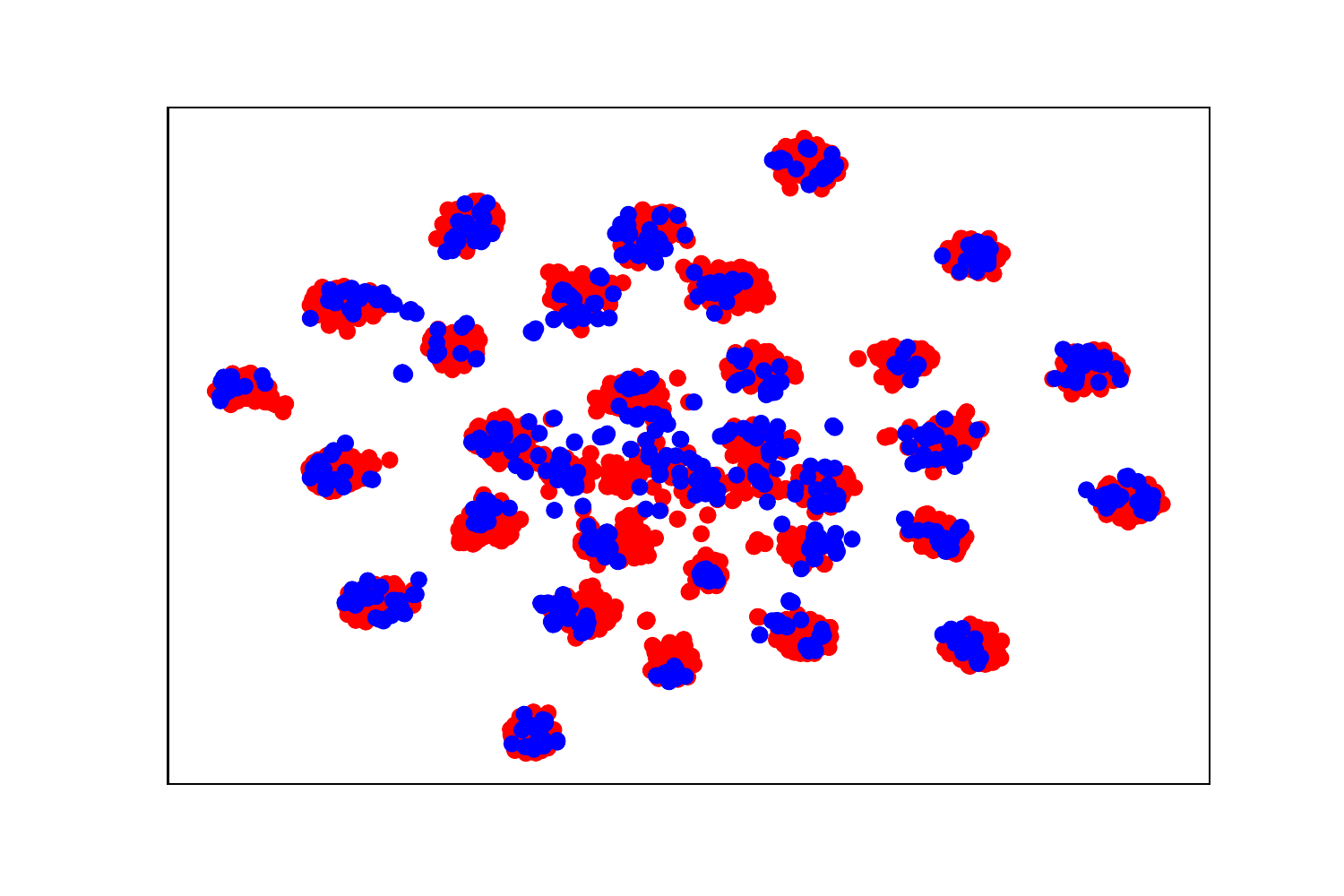}}
    \hspace{-5mm}
    \subfigure[CSCAL]{\includegraphics[width=0.32\linewidth]{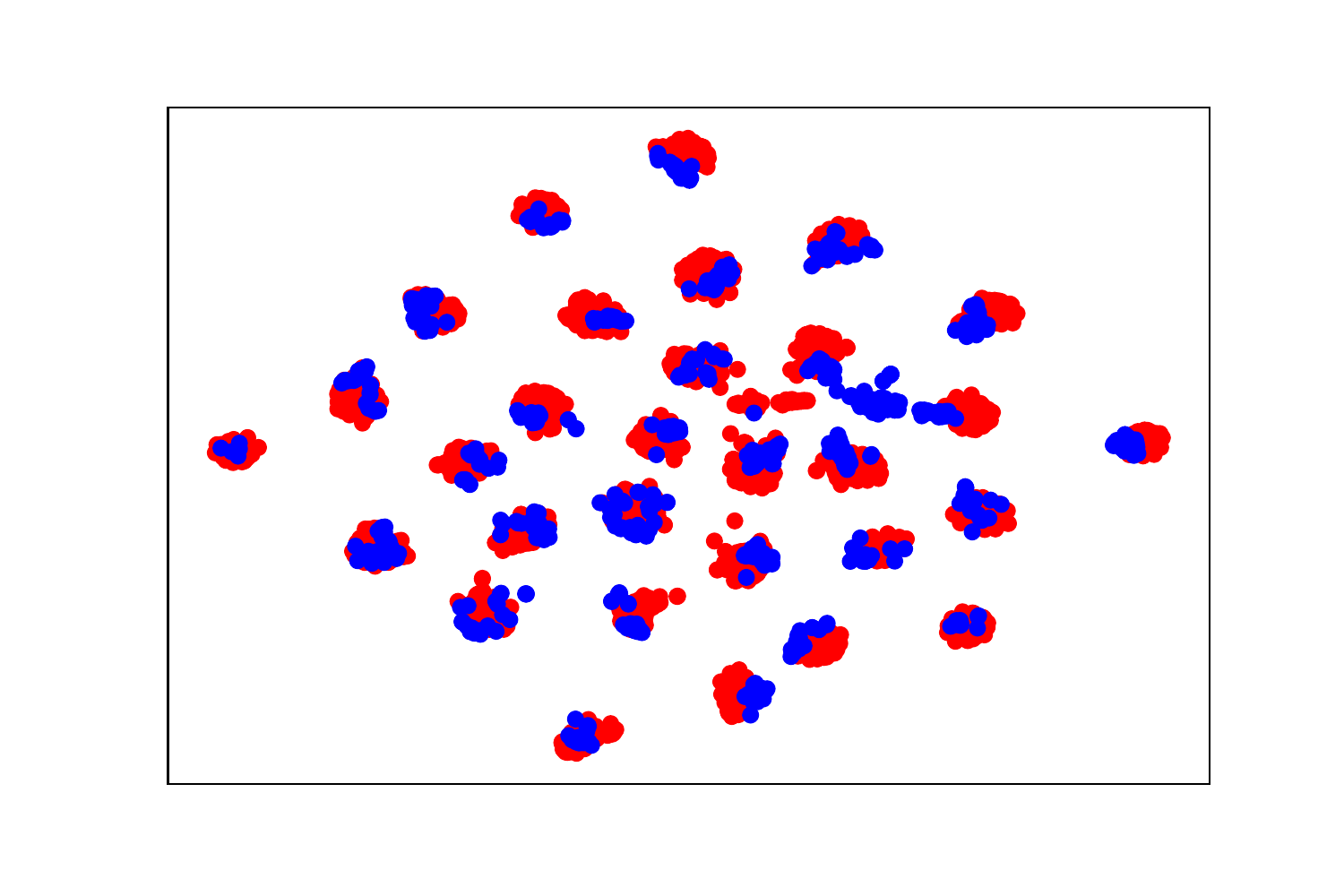}}
    \hspace{-5mm}
    \subfigure[DANN+CSCAL]{\includegraphics[width=0.32\linewidth]{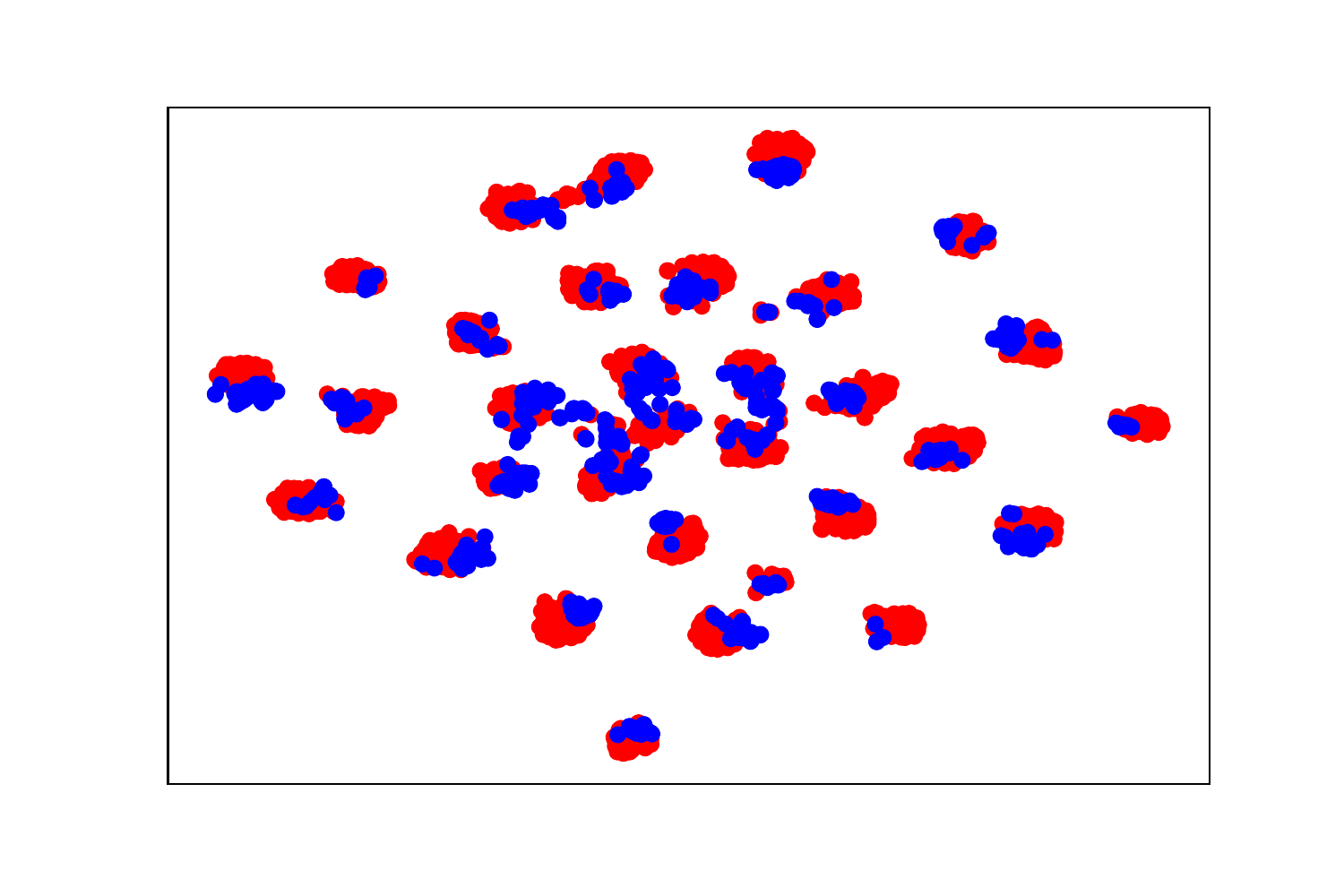}}
    \hspace{-5mm}
    \subfigure[JANN+CSCAL]{\includegraphics[width=0.32\linewidth]{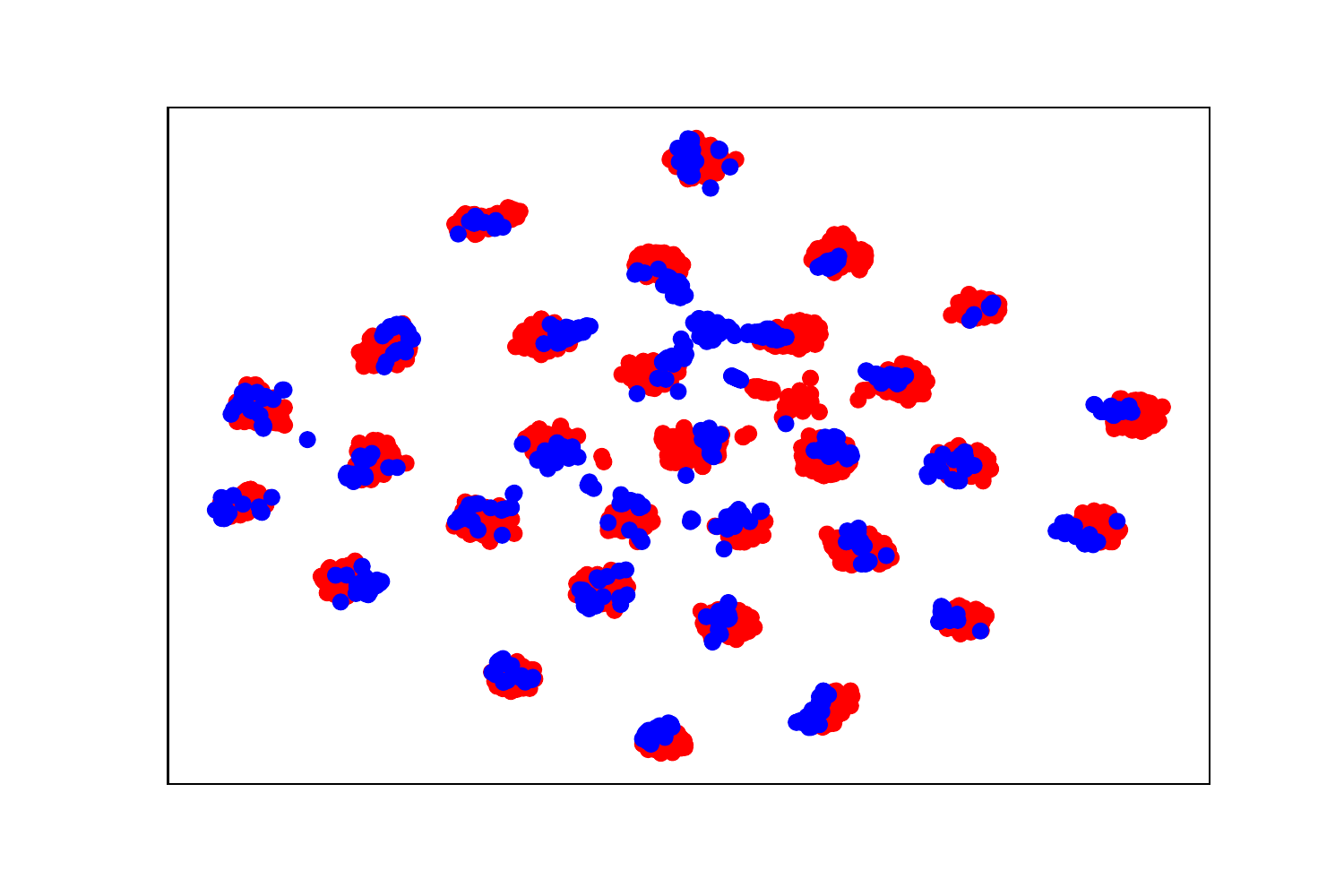}}
    \caption{The t-SNE of different methods on task A $\rightarrow$ W of Office-31. Source and target features are represented by red and blue scatter points, respectively.}
    \label{fig:fig5}
\end{figure}
\begin{figure}[!t]
    \centering
    \subfigcapskip=-5pt
    \subfigure[A$\rightarrow$ W]{\includegraphics[width=0.32\linewidth]{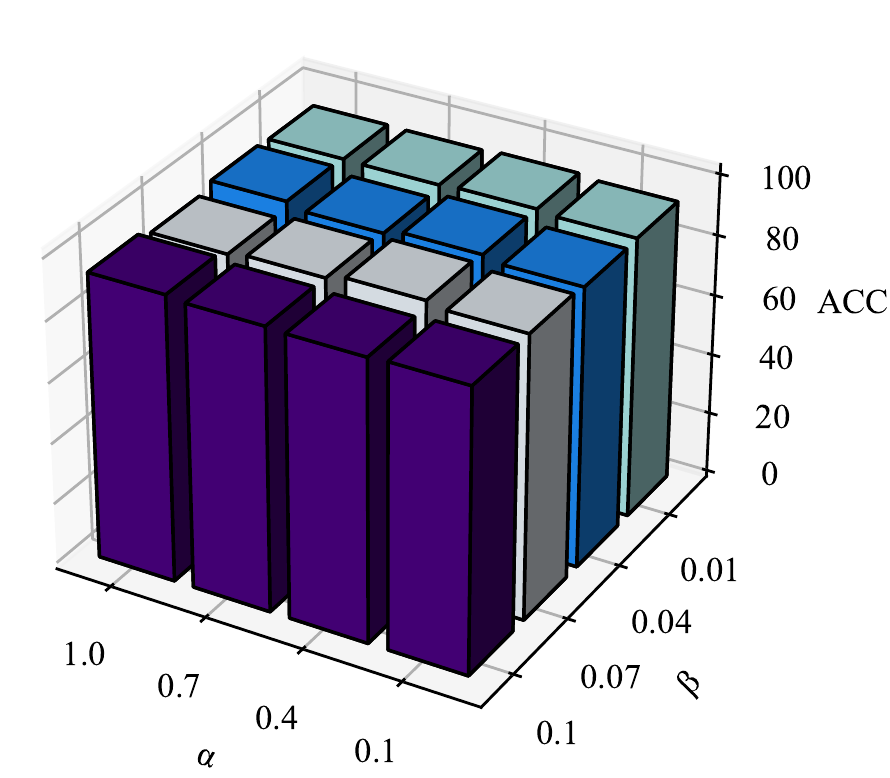}}
    \subfigure[A$\rightarrow$D]{\includegraphics[width=0.32\linewidth]{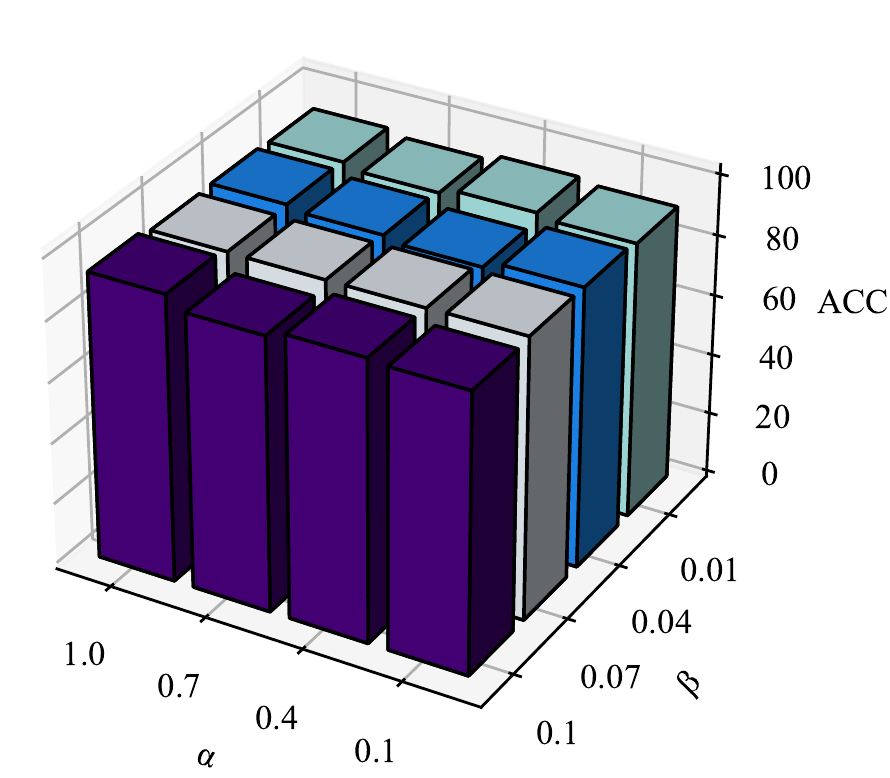}}
    \subfigure[D $\rightarrow$ W]{\includegraphics[width=0.32\linewidth]{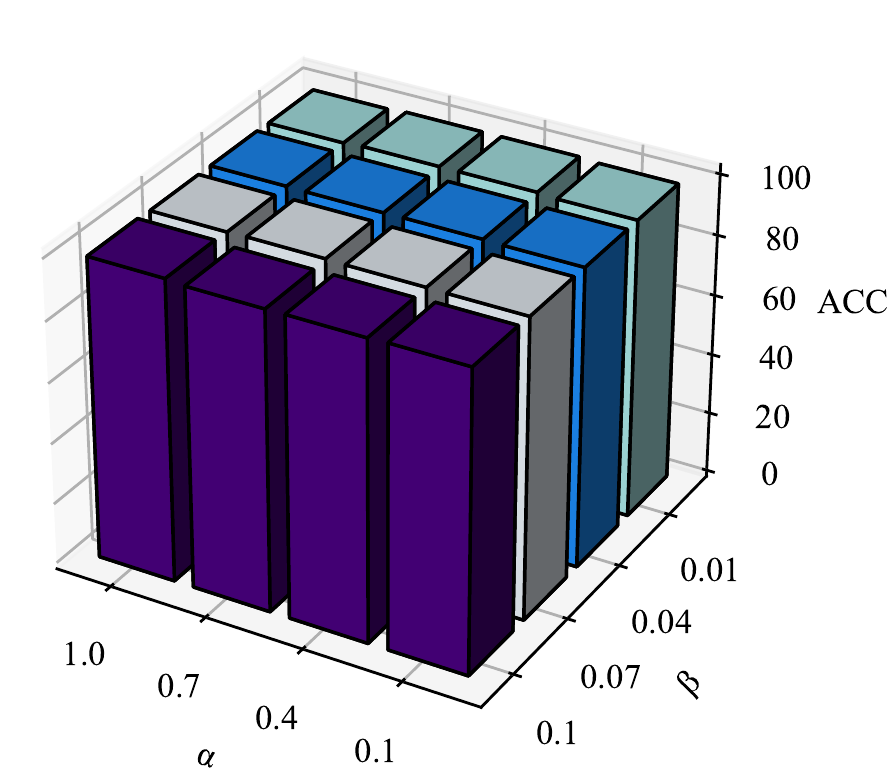}}
    \caption{The visualization of CSCAL sensitivity to parameters $\alpha$ and $\beta$ on tasks A$\rightarrow$ W, A$\rightarrow$ D and D$\rightarrow$ W, respectively}
    \label{fig:fig6}
\end{figure}

\textbf{t-SNE Visualization.} The t-SNE visualization~\cite{van2008visualizing} of learned feature representations of different methods is shown in Fig.~\ref{fig:fig5}.
The ResNet-50, without any adaptation, has the poorest performance in aligning source and target features. 
Compared with DANN and JANN, CSCAL mitigates the discrepancy between the source and target domain well and has more discriminative features across different categories.
Additionally, compared with the original methods, DANN+CSCAL and JANN+CSCAL generate more domain alignment and category distinguishment results.

\textbf{Parameter Sensitivity.}
The Fig.~\ref{fig:fig6}. visualizes the sensitivity of CSCAL about the hyper-parameters $\alpha$ and $\beta$ in loss function Eq.~\ref{eq9} on tasks A$ \rightarrow $W, A$ \rightarrow $D, and W$ \rightarrow $D when $\alpha \in \{1.0, 0.7, 0.4, 0.1\}$ and $\beta \in \{0.1, 0.07, 0.04, 0.01\}$.
The results show that CSCAL is not sensitive to hyper-parameters $\alpha$ and $\beta$.

\begin{table}[htbp]
    \centering
    \vspace{-0.2cm} 
    \setlength{\tabcolsep}{4pt}
    \caption{Classification Accuracy (\%) on Office-31.
    $^\dag$ denotes that results are reproduced according to the public source code.}
    \begin{tabular}{|l|c c c c c c c|}
    \hline
        Method & A$\rightarrow$W & D$\rightarrow$W & W$\rightarrow$D & A$\rightarrow$D & D$\rightarrow$A & W$\rightarrow$A & Avg\\
    \hline
    \hline
         ResNet-50~\cite{he2016deep}& 68.4 & 96.7 & 99.3 & 68.9 & 62.5 & 60.7 & 76.1 \\ WDGRL~\cite{shen2018wasserstein} & 72.6 & 97.1 & 99.2 & 79.5 & 63.7 & 59.5 & 78.6 \\ MCD~\cite{saito2018maximum} & 88.6 & 98.5 & \textcolor{red}{\textbf{100.0}} & 92.2 & 69.5 & 69.7 & 86.5 \\ 
         BNM~\cite{cui2020towards} & 91.5 & 98.5 & \textcolor{red}{\textbf{100.0}} & 90.3 & 70.9 & 71.6 & 87.1 \\
         ETD~\cite{li2020enhanced} & 92.1 & \textcolor{red}{\textbf{100.0}} & \textcolor{red}{\textbf{100.0}} & 88.0 & 71.0 & 67.8 & 86.2 \\
         SymNets~\cite{zhang2019domain} & 90.8 & 98.8 & \textcolor{red}{\textbf{100.0}} & 93.9 & 74.6 & 72.5 & 88.4 \\
         TSA~\cite{li2021transferable} & \textcolor{blue}{94.8} & \textcolor{blue}{99.1} & \textcolor{red}{\textbf{100.0}} & 92.6 & 74.9 & 74.4 & 89.3 \\ SCDA$^{\dag}$~\cite{li2021semantic} & 93.9 & 98.6 & \textcolor{red}{\textbf{100.0}} & 94.2 & \textcolor{blue}{75.6} & \textcolor{red}{\textbf{76.0}} & \textcolor{blue}{89.7}\\
         SUDA~\cite{zhang2022spectral} & 90.8 & 98.7 & \textcolor{red}{\textbf{100.0}} & 91.2 & 72.2 & 71.4 & 87.4 \\
         \hline
         CSCAL & 94.6 & \textcolor{blue}{99.1} & \textcolor{red}{\textbf{100.0}} & 94.2 & \textcolor{red}{\textbf{76.5}} & \textcolor{blue}{75.6} & \textcolor{red}{\textbf{90.0}}\\
         \hline
         DANN~\cite{ganin2015unsupervised} & 82.0 & 96.9 & 99.1 & 79.7 & 68.2 & 67.4 & 82.2 \\
         DANN+CSCAL & 90.8 & 98.7 & \textcolor{red}{\textbf{100.0}} & 88.6 & 69.8 & 67.7 & 85.9\\
         \hline
         JANN~\cite{long2017deep} & 85.4 & 97.4 & 99.8 & 84.7 & 68.6 & 70.0 & 84.3\\
         JANN+CSCAL & 93.5 & 97.6 & \textcolor{red}{\textbf{100.0}} & 90.0 & 73.0 & 71.6 & 87.6\\
         \hline
         MCC~\cite{jin2020minimum} & \textcolor{red}{\textbf{95.5}} & 98.6 & \textcolor{red}{\textbf{100.0}} & \textcolor{red}{\textbf{94.4}} & 72.9 & 74.9 & 89.4 \\
         MCC+CSCAL & \textcolor{blue}{94.8} & 98.4 & 99.8 & \textcolor{red}{\textbf{94.4}} & 74.4 & 74.9 & 89.5\\
         \hline
    \end{tabular}
    \label{tab:tab3}
\end{table}

\begin{table}[htbp]
    \caption{Ablation Study of CSCAL on Office-31 (ResNet-50)}
    \centering
    \setlength{\tabcolsep}{2pt}
    \begin{tabular}{|l|c c c c c c c|}
    \hline
        Method & A$\rightarrow$W & D$\rightarrow$W & W$\rightarrow$D & A$\rightarrow$D & D$\rightarrow$A & W$\rightarrow$A & Avg\\
    \hline
    \hline
        ResNet-50& 68.4 & 96.7 & 99.3 & 68.9 & 62.5 & 60.7 & 76.1\\
        + CSCAL(w/o $\mathcal{L}_{pld}$) & 92.0 & 98.4 &100.0 & 93.6 & 74.7 & 75.2 & 89.0\\
        + CSCAL(w/o $\mathcal{L}_{nnd}$) & 92.6 & 98.5 & 100.0 & 94.0 & 75.5 & 75.7 & 89.4\\
        + CSCAL(w/o $\mathcal{L}_{mi}$) & 90.1 & 98.7 & 100.0 & 91.8 & 75.4 & 65.7 & 87.0 \\
        + CSCAL & 94.6 & 99.1 & 100.0 & 94.2 & 76.5 & 75.6 & 90.0 \\
    \hline
    \end{tabular}
    \label{tab:tab4}
\end{table}

\section{Conclusions}
In this paper, we proposed a novel classifier-based adversarial paradigm, CSCAL, which focuses on crucial semantic transferring without an additional well-designed domain discriminator.
Specifically, we designed PLD to discover and adapt the crucial semantic knowledge via an intra-class-wise adversarial alignment.
Further, inter-class-wise information is considered in NND construction and further model adaptation.
Moreover, CSCAL can dramatically improve various UDA methods' performance.
Extensive experimental results demonstrate the CSCAL effectiveness.

\bibliography{reference}
\end{document}